\documentclass[lettersize,journal]{IEEEtran}
\usepackage{amsmath,amsfonts}
\usepackage{array}
\usepackage[caption=false,font=normalsize,labelfont=sf,textfont=sf]{subfig}
\usepackage{textcomp}
\usepackage{stfloats}
\usepackage{url}
\usepackage{verbatim}
\usepackage{graphicx}
\usepackage{cite}
\usepackage{bbm} 
\usepackage{multirow}
\usepackage{booktabs}
\usepackage{amssymb}
\usepackage{xcolor}
\usepackage{bibunits}

\usepackage[linesnumbered,ruled,vlined]{algorithm2e}

\SetCommentSty{mycommfont}
\SetKwInput{KwInput}{Input}                
\SetKwInput{KwOutput}{Output}              
\SetKwInput{KwNT}{Network Training}              

\begin{document}

\title{IRASNet: Improved Feature-Level Clutter Reduction for Domain Generalized SAR-ATR}
\author{Oh-Tae Jang, Min-Jun Kim, Sung-Ho Kim, Hee-Sub Shin, and Kyung-Tae Kim,~\IEEEmembership{Member,~IEEE}

\thanks{This work was supported by Korea Research Institute for defense Technology planning and advancement (KRIT) - Grant funded by Defense Acquisition Program Administration (DAPA) (KRIT-CT-22-060).  \textit{(Corresponding author: Kyung-Tae Kim.)}}
\thanks{Oh-Tae Jang, and Kyung-Tae Kim are with the Department of Electrical Engineering, Pohang University of Science and Technology, Pohang 790-784, South Korea (e-mail: otaejang@postech.ac.kr; kkt@postech.ac.kr).}
\thanks{Min-Jun Kim, and Sung-Ho Kim are with the Department of Electronic Engineering, Yeungnam University, Gyeongsan 71-38, South Korea (e-mail: kmj17211@yu.ac.kr; sunghokim@yu.ac.kr).}
\thanks{Hee-Sub Shin is with LIG Nex1 Co., Ltd., South Korea (e-mail: heesub.shin2@lignex1.com).}}

\IEEEpubid{    }

\maketitle

\begin{abstract}
Recently, computer-aided design models and electromagnetic simulations have been used to augment synthetic aperture radar (SAR) data for deep learning. However, an automatic target recognition (ATR) model struggles with domain shift when using synthetic data because the model learns specific clutter patterns present in such data, which disturbs performance when applied to measured data with different clutter distributions. This study proposes a framework particularly designed for domain-generalized SAR-ATR called IRASNet, enabling effective feature-level clutter reduction and domain-invariant feature learning. First, we propose a clutter reduction module (CRM) that maximizes the signal-to-clutter ratio on feature maps. The module reduces the impact of clutter at the feature level while preserving target and shadow information, thereby improving ATR performance. Second, we integrate adversarial learning with CRM to extract clutter-reduced domain-invariant features. The integration bridges the gap between synthetic and measured datasets without requiring measured data during training. Third, we improve feature extraction from target and shadow regions by implementing a positional supervision task using mask ground truth encoding. The improvement enhances the ability of the model to discriminate between classes. Our proposed IRASNet presents new state-of-the-art public SAR datasets utilizing target and shadow information to achieve superior performance across various test conditions. IRASNet not only enhances generalization performance but also significantly improves feature-level clutter reduction, making it a valuable advancement in the field of radar image pattern recognition.
\end{abstract}

\section{Introduction}
\IEEEPARstart{I}{n remote} sensing, synthetic aperture radar (SAR) is primarily used as a radar imaging system utilizing a wide range of frequencies to generate high-resolution images. Compared to optical images, SAR plays an important role in surveillance and reconnaissance because of its ability to produce images under all-day, all-weather, and long-range conditions \cite{cumming2005digital}. Conversely, SAR images of objects under interest show complicated electromagnetic scattering phenomena, including various scattering mechanisms owing to substructures on targets, clutter or interfering signals, and speckle noise. Unlike optical images, SAR images hold intensity and phase information, while lacking color information. Therefore, a visual understanding of SAR images is quite challenging \cite{choi2022fusion,leefilter,frost1982model}. Accordingly, manual analysis of large SAR image streams requires a significant amount of human resources, which has led to the development of SAR automatic target recognition (ATR). Traditional SAR-ATR algorithms are focused on building handcrafted features and an appropriate classifier for SAR images \cite{globalandlocalfeature,thiagarajan2010sparse,zhang2012multi,potter1997attributed}.

Owing to advances in deep learning (DL), several SAR-ATR approaches based on convolutional neural networks (CNNs) have been proposed to automatically extract optimal features from input data \cite{wilmanski2016modern,yu2021lightweight,chen2016target,amcnn,ying2020tai}. DL-based SAR-ATR algorithms significantly outperform traditional approaches based on handcrafted features, owing to their abilities based on optimal feature extraction. However, huge amounts of datasets are required to guarantee reliable ATR performance \cite{lecun2015deep}. Unfortunately, the process of collecting and labeling measured SAR images to train an ATR model is highly time-consuming and expensive; therefore, obtaining a sufficient number of datasets for DL-based SAR-ATR is challenging \cite{zhang2023vsfa}.

\IEEEpubidadjcol
One effective solution to address the limited availability of SAR data utilizes synthesized datasets generated by numerical electromagnetic solvers with realistic computer-aided design (CAD) models for targets of interest \cite{mocem,zhang2022sar, hammer2009coherent, balz2009hybrid}. For instance, the synthetic and measured paired labeled experiment (SAMPLE) dataset \cite{smpl} can be employed. The approach generates numerous synthetic SAR images within a reasonable period, containing physically meaningful and reliable backscattered echoes from targets of interest \cite{zhang2023vsfa}. Data augmentation using CAD models and electromagnetic simulations mitigates overfitting in DL models, making the approach particularly effective in SAR-ATR, where data are often limited. 

However, even when care is taken to exactly match actual physical situations, errors may still arise owing to geometric inaccuracies in CAD models \cite{malmgren2017improving, smpl,moore2010analytical} and approximations during numerical calculations \cite{hammer2009coherent, balz2009hybrid}, leading to an imperfect representation of real-world phenomena. Such discrepancies cause a domain shift, manifested as a difference in the distribution between synthetic and measured SAR datasets. Consequently, classifiers trained on only synthetic datasets tend to exhibit reduced accuracy when applied to measured datasets \cite{smpl}. Therefore, bridging the domain gap and accurately classifying measured data using synthetic data has become a popular research focus.

Most researchers have attempted to address the main issue of domain shift through domain adaptation (DA) research \cite{han2024kd, tdda, sun_cie_radar, cross_domain, chen2022pixel,sun2023gradual}. DA mitigates domain shift in feature spaces by extracting domain-invariant features through learning on a dataset that simultaneously includes both synthetic and measured data, thereby improving ATR performance \cite{dan,dann}. Nonetheless, DA relies on a strong assumption that the measured data are accessible for model adaptation, which is not always feasible in real-world operating conditions \cite{zhou2022domain}. For example, obtaining measured SAR images of specific targets may not always be possible in every training scenario due to the cost and limited availability of real targets required for data collection. Additionally, since intelligence, surveillance, and reconnaissance  SAR are closely related to military purposes, access to measured data is often restricted due to security concerns. Such constraints make it unrealistic to always use measured data for adaptation. Therefore, the applicability of DA is limited in numerous scenarios.

Conversely, DL-based SAR-ATR research currently focuses on domain generalization (DG), also known as out-of-distribution generalization \cite{blanchard2011generalizing} to overcome the domain shift problem and the absence of measured data \cite{inkawhich2021bridging, mocem, zhang2022sar, mj}. Unlike DA, DG relies solely on synthetic datasets for training, excluding measured datasets \cite{zhou2022domain, Li_2018_CVPR, li2018deep}. The definition of DG addresses a more realistic ``open world" assumption, where not all targets of interest encountered in a field can be assumed to be learned \cite{inkawhich2021bridging}. The strategy is particularly suitable for SAR-ATR problems where access to measured data is challenging.

Current approaches to address domain-generalized SAR-ATR (DG-ATR) problems aim to overcome domain gap through data augmentation methods \cite{inkawhich2021bridging, mj} and hierarchical recognition methods based on multi-similarity fusion (MSF) \cite{zhang2022sar}. Data augmentation methods attempt to solve DG-ATR problems by diversifying the distribution of synthetic datasets, particularly by adding noise to increase distributional variety \cite{inkawhich2021bridging, mj}. Nevertheless, such approaches raise concerns about the extent of noise required to achieve sufficient distributional diversity owing to their excessive reliance on augmented data \cite{chen2022pixel}. Moreover, from a DG perspective, data augmentation aims to reduce the domain gap at a pixel level and does not necessarily ensure its reduction at a feature level. However, MSF and hierarchical recognition techniques \cite{zhang2022sar} also exhibit limitations. The methods encounter consistency issues when images with different distributions are introduced at the target/shadow segmentation phase in preprocessing. Conversely, segmentation algorithms experience domain variance, and pixel-level discrepancies between domains can lead to larger errors at the feature level.

\begin{figure}[t]
\centering
\includegraphics[width=0.9\columnwidth]{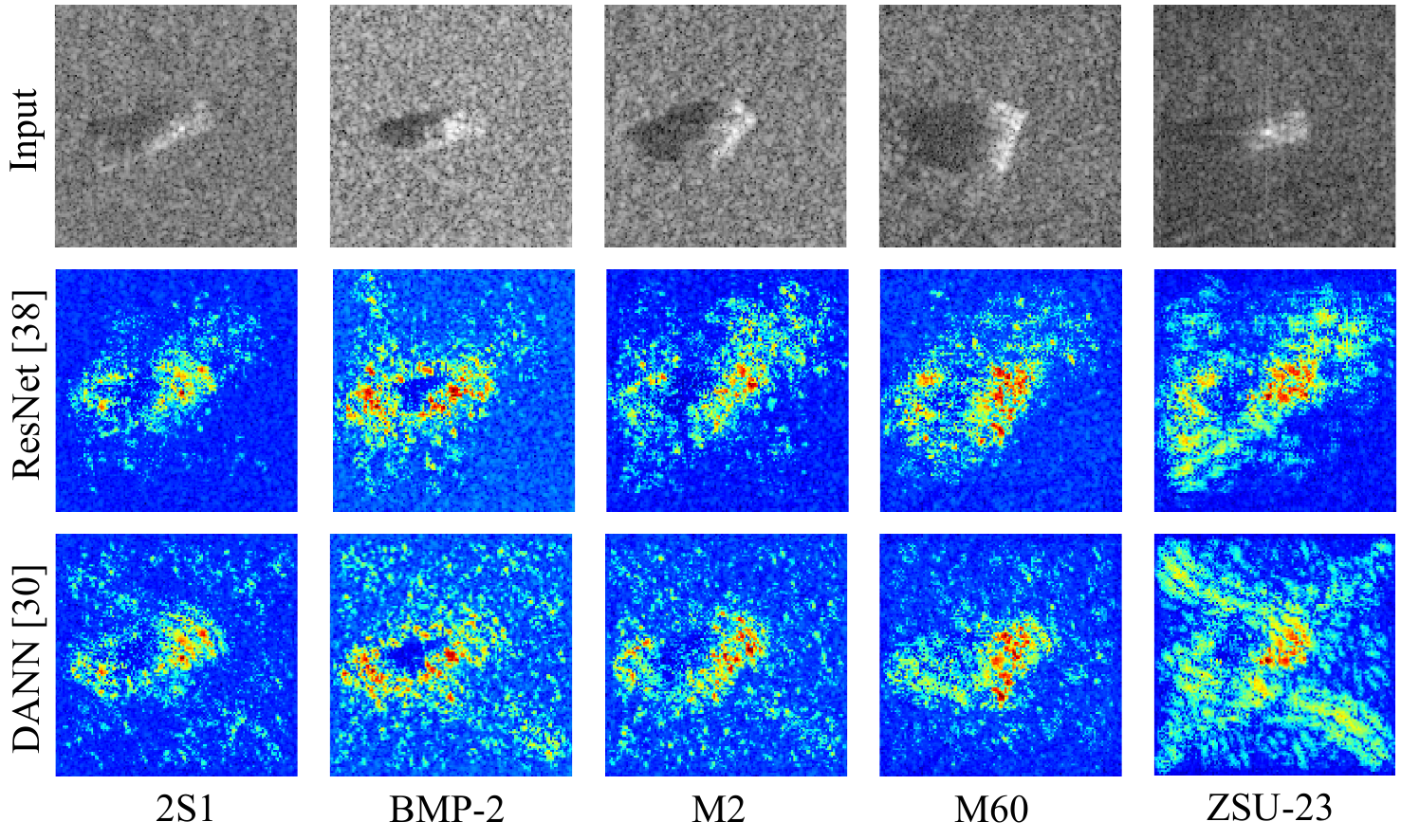}
\caption{Visualization of input image contributions in DL models using Shapley additive explanations (SHAP) \cite{lundberg2017unified} for ResNet \cite{resnet} and DANN \cite{dann}.}
\vspace{-1em}
\label{fig:shap}
\end{figure}

To address existing challenges, a novel DG framework for SAR-ATR named integrating clutter reduction module (CRM) and adversarial learning (IRASNet) is proposed in this study. As shown in Fig.~\ref{fig:shap}, extracting domain-invariant features solely through adversarial learning can lead to overfitting in clutter regions owing to the inherent challenge of learning clutter. Additionally, separating the target and shadow regions during preprocessing can cause underfitting, stemming from the high domain variation associated with segmentation techniques. Unlike existing algorithms, IRASNet addresses both of the challenges simultaneously by combining clutter reduction at the feature level through a CRM and extracting domain-invariant features through adversarial learning to achieve DG. Additionally, incorporating shadow features into clutter-reduced features can improve ATR performance \cite{choi2022fusion}. Moreover, higher performance can be anticipated by increasing SCR at the feature level to focus on the target region.

The main contributions in this paper are summarized as follows:

\begin{enumerate}
    \item A novel DG framework for SAR-ATR named IRASNet is proposed. To the best of our knowledge, this is the first study to report integrating feature-level clutter reduction and domain-invariant feature learning within a network to improve DG SAR-ATR. Furthermore, our method achieves state-of-the-art results on a public SAMPLE dataset in terms of DG.
    \item This paper proposes CRM instead of target and shadow segmentation during preprocessing, which reduces processing time and enables target and shadow-based processing in the feature space. Specifically, the module is capable of accurately classifying images even when the SCR of the test image is low and the background clutter image is changed.
    \item This paper proposes a simple yet effective positional supervision task. The introduction of mask ground truth (Mask GT) encoding of targets and shadows helps to reliably reflect the position information of targets and shadows for feature-level clutter reduction.
\end{enumerate}


\begin{table*}[t]
\centering
\caption{Performance of DANN and ResNet trained on synthetic data and tested across different domains \\ with different data preprocessing techniques (for example, pixel-level clutter reduction).}
\label{tab:ori_plcr}
\renewcommand{\arraystretch}{1.6} 
\resizebox{0.8\textwidth}{!}{
\begin{tabular}{@{}ccclcclccl@{}}
\toprule
\toprule
\multirow{2}{*}{} & \multicolumn{2}{c}{\textbf{Synthetic}}                                                                                                     &  & \multicolumn{2}{c}{\textbf{Measured}}                                                                                                      &  & \multicolumn{2}{c}{\textbf{Unknown Clutter}}                                                                                                &  \\ \cmidrule(l){2-10} 
                  & \begin{tabular}[c]{@{}c@{}}Original\\ Img\end{tabular} & \begin{tabular}[c]{@{}c@{}}Target + Shadow\\ (Segmentation)\end{tabular} &  & \begin{tabular}[c]{@{}c@{}}Original\\ Img\end{tabular} & \begin{tabular}[c]{@{}c@{}}Target + Shadow\\ (Segmentation)\end{tabular} &  & \begin{tabular}[c]{@{}c@{}}Original\\ Img\end{tabular} & \begin{tabular}[c]{@{}c@{}}Target + Shadow\\ (Segmentation)\end{tabular} &  \\ \midrule
\quad DANN\cite{dann}             & 99.89\%                                                & 99.91\%                                                                  &  & 91.30\%                                                & 70.56\%                                                                  &  & 89.22\%                                                & 62.45\%                                                                  &  \\
\quad ResNet\cite{resnet}           & 99.92\%                                                & 99.84\%                                                                  &  & 93.76\%                                                & 70.11\%                                                                  &  & 87.73\%                                                & 58.81\%                                                                  &  \\ \bottomrule
\end{tabular}%
}
\end{table*}

\section{Related Works}
In this section, the vulnerability of current domain-invariant representation learning (DIRL)-based SAR-ATR approaches \cite{chen2022pixel, dann} in handling clutter areas is experimentally demonstrated. Subsequently, we explore solutions for successful DG-ATR.

\begin{figure}[t]
\centering
\includegraphics[width=\columnwidth]{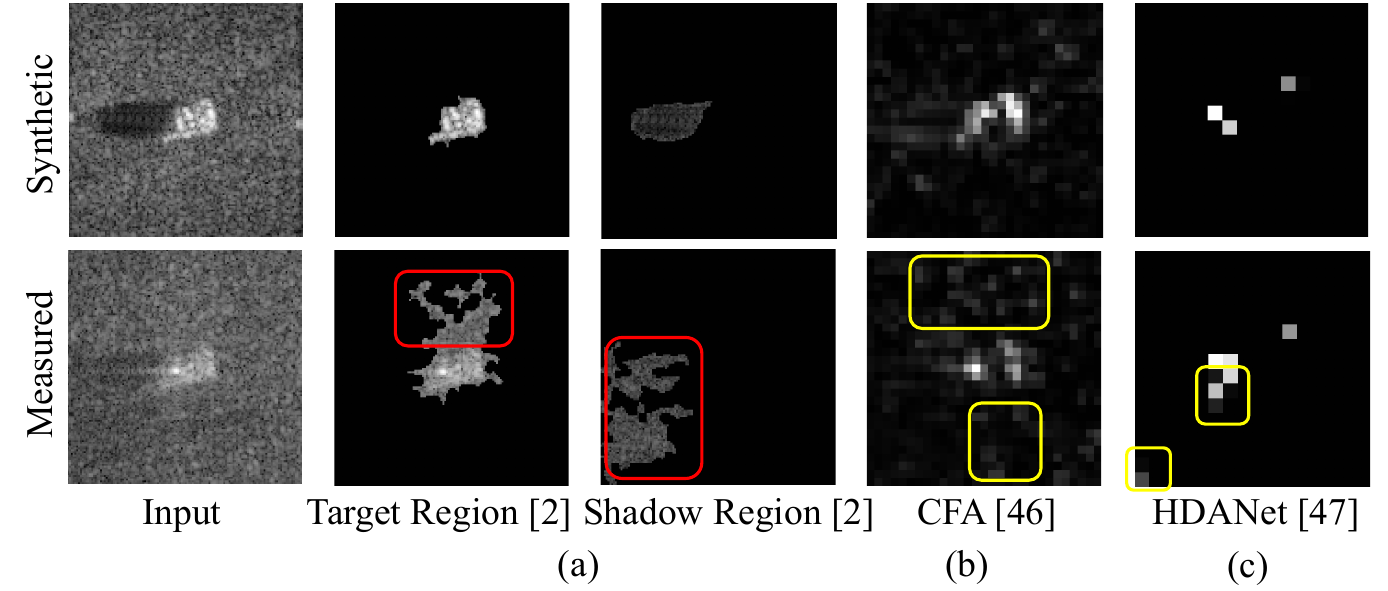}
\caption{Visualization of algorithms for reducing the impact of clutter according to domain. (a) Pixel-level clutter reduction algorithm. (b) Clutter-robust learning algorithm. (c) Feature-level clutter reduction algorithm. The \textit{\textbf{\textcolor{red}{red}}} bounding box shows pixel-level errors, and the \textit{\textbf{\textcolor{yellow}{yellow}}} bounding box shows feature-level errors.}
\label{fig:segerror_featureerror}
\end{figure}

\begin{figure*}[t]
\centering
\includegraphics[width=\linewidth]{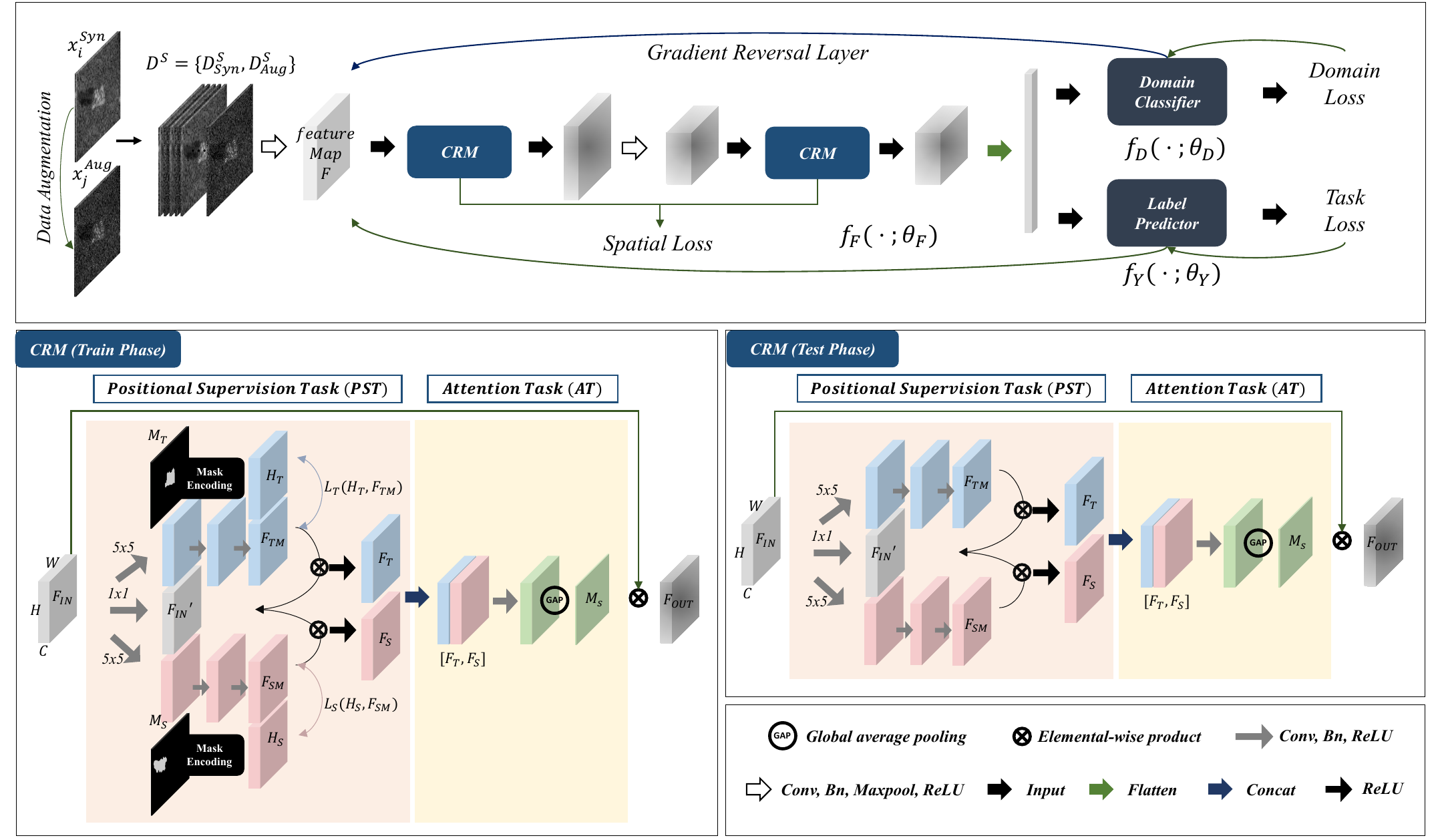}
\caption{Overall pipeline of the proposed IRASNet framework. The CRM operates differently during the training and testing phases. The augmented and synthetic datasets are combined using a novel CRM and adversarial learning to derive domain-invariant, clutter-reduced features.}
\vspace{-1em}
\label{fig:irasnet}
\end{figure*}

General DL frameworks are prone to learning clutter regions. Belloni et al. \cite{belloni2020explainability}, Li et al. \cite{li2023discovering}, and Heiligers et al. \cite{heiligers2018importance} demonstrated that data and model biases caused models to exhibit non-causality to background clutter, leading to overfitting and undesirable ATR behavior. To analyze their finding in the context of DG, explainable artificial intelligence (XAI), specifically Shapley additive explanations (SHAP) \cite{lundberg2017unified}, is employed to directly visualize the rationale behind decisions of the network, as illustrated in Fig.~\ref{fig:shap}. SHAP allows for the visual inspection of how each pixel in an input image impacts the prediction of a DL model. The DL model involves a network for DG based on adversarial learning \cite{dann}, trained on synthetic and noisy data from the SAMPLE dataset \cite{smpl}, and tested on five classes of measured data. Despite overcoming domain shifts and extracting domain-invariant features through DIRL, the results show that the network makes decisions by referencing clutter regions. Therefore, clutter is a significant obstacle to DG-ATR problems in general DL frameworks. Thus, models require training to better learn true causal features, specifically the intrinsic characteristics of their targets, while relying less on background clutter.

Several algorithms have been proposed to help networks learn the intrinsic characteristics of their targets more effectively by removing clutter \cite{zhou2018sar,zhang2022sar,wang2019sar, choi2022fusion}. Pixel-level clutter reduction methods ensure high performance in ATR problems with the same distribution of train and test data, as the segmentation mask remains consistent. Conversely, applying DG-ATR encounters limitations in ensuring domain invariance during the preprocessing of targets and shadow segmentations due to different hyper-parameters for different domains. Segmentation algorithms for pixel-level clutter reduction require hyper-parameters for specifying candidate regions of the target and shadow areas, and the SCR between the target and clutter \cite{meth1998target, zhou2018sar}. Therefore, such methods do not work accurately when the candidate regions for the target and shadow are inconsistent or when a sufficient SCR is not maintained. 

As illustrated in Fig.~\ref{fig:segerror_featureerror} (a), when the SCR is low in the measured image, the information in the target and shadow region can be entirely lost, despite applying the same segmentation method. The loss cannot be compensated within the DL network \cite{strang2012linear}, leading to degraded performance. As indicated in Table~\ref{tab:ori_plcr}, performance significantly declines when training and testing with only target and shadow regions compared to using images that include clutter regions. Even in the DIRL performed by DANN \cite{dann}, the domain gap was not bridged, likely due to information loss during segmentation. Therefore, performing feature-level clutter reduction to avoid information loss is imperative, even if it results in less effective reduction of clutter regions. 

On the other hand, several methodologies have focused on reducing the impact of clutter at the feature level. Peng et al. \cite{cfa} proposed contrastive feature alignment (CFA) that combined a mixed clutter variant generation strategy with channel-weighted mean square error (CWMSE) loss to extract robust features against clutter. The approach effectively enhanced target recognition robustness in various background clutter scenarios. However, the method posed issues of still retaining clutter values at the feature level, which could lead to reliance on clutter for decision-making and poor generalization performance. Therefore, methods that learn to suppress clutter and reduce its influence at the feature level are essential. 

Furthermore, Li et al. \cite{li_HDANet} proposed a hierarchical disentanglement-alignment network (HDANet) that reduced clutter and extracted target features through multi-task-assisted mask disentanglement and domain alignment. However, the approach did not utilize shadow information at the feature level, resulting in a decrease in ATR performance compared to when both target and shadow information were used together \cite{choi2022fusion}. Herein, Fig.~\ref{fig:segerror_featureerror} (b) and (c) present the features of synthetic and measured data for CFA\cite{cfa} and HDANet\cite{li_HDANet} trained on synthetic and augmented datasets, respectively. On examining the yellow bounding boxes, high values remained in the clutter region of the measured data, indicating that clutter reduction was not perfectly achieved. Currently, researchers have not extensively studied clutter reduction in the context of DG. To enhance DG performance, ensuring both domain invariance and clutter reduction simultaneously is essential.

\section{Methodology}
IRASNet is a framework designed to address the DG problem, achieving high performance on measured data when trained solely on a $100\%$ synthetic dataset. The overall concept of the proposed IRASNet is presented in Fig.~\ref{fig:irasnet}. This section details the stepwise procedures of the proposed framework. 

\subsection{Problem Formulation}
We introduce the definition of the DG problem. We denote the input and label spaces by \(\mathcal{X}\) and \(\mathcal{Y}\), respectively. The dataset \(X\) comprises data points extracted from the input space \(\mathcal{X}\), and the corresponding label set \(Y\) is drawn from the label space \(\mathcal{Y}\). Therefore, \(X \subset \mathcal{X}\) and \(Y \subset \mathcal{Y}\) represent the subsets of their respective spaces. The datasets \(X\) and \(Y\) are structured as: \(X = \{x_1, x_2, \dots, x_i\}\) and \(Y = \{y_1, y_2, \dots, y_i\}\), where each \(x_i\) and \(y_i\) represent an image (or input data) and the corresponding label, respectively. 

In the context of DG, we assume to have two domains: \( K \) similar but distinct source domains \( D^S = \{D^S_k = \{X^{(k)},Y^{(k)}\}\}_{k=1}^K \), where each source domain is defined according to the previously defined domain with joint distribution \( P_{XY}^{(k)} \) and the target domain \( D^T \). Notably, \( P_{XY}^{(k')} \neq P_{XY}^{(k)} \) and \( k \neq k' \) with \( k, k' \in \{1, \ldots, K\} \). DG aims to learn a predictive model \( f: \mathcal{X} \rightarrow \mathcal{Y} \) using only source domain data such that the prediction error on an unseen target domain \( D^T = \{X^T\} \). The target domain is not used at all during training; therefore, the label space is not defined. Naturally, the joint distribution of the target domain \( D^T \) is denoted as \( P_{XY}^T \), and \( P_{XY}^T \neq P_{XY}^{(k)} \) $\forall$ \( k \in \{1, \ldots, K\} \).

\subsection{Building Multiple Source Domain}
The proposed framework utilizes data augmentation from synthetic datasets based on existing studies \cite{mj, inkawhich2021bridging} for robust clutter reduction and constructing multiple source domains. We obtain an augmented domain \(D_{Aug}^S\) through augmentation applied to the synthetic domain \(D_{Syn}^S\). Thus, the constructed source domain is defined as two similar but distinct source domains \(D^S = \{D_{Syn}^S, D_{Aug}^S\}\), where \(P_{XY}^{Syn} \neq P_{XY}^{Aug}\). Herein, \(D_{Syn}^S = \{(X^{Syn}, Y^{Syn}, M^{Syn}_T, M^{Syn}_S)\}\) represents the synthetic domain, where the mask set \(M_{T/S}\), defined as target/shadow mask set for training in IRASNet, is included in the source domain. Conversely, the mask set \(M_{T/S}\) corresponding to \(X\) is drawn from the mask space \(\mathcal{M}\) (\(M_{T/S} \subset \mathcal{M}\)). Additionally, the datasets \(M_{T/S}\) are structured as: \(M_{T/S} = \{m_{t/s,1}, \dots, m_{t/s,i}\}\), where each \(m_{t/s,i}\) represent a target/shadow mask. Furthermore, \(D_{Aug}^S = \{X^{Aug}, Y^{Aug}, M^{Aug}_T, M^{Aug}_S\}\) represents the augmented domain derived from the synthetic dataset. The measured dataset is defined as the target domain, specifically represented as \(D^T= \{D_{Mea}^T\}\), where \(D_{Mea}^T = \{X^{Mea}\}\) is the measured domain. Notably, the measured dataset does not include labels or a mask set. The detailed methods and hyperparameters used for augmentation and mask generation are provided in the \textit{supplementary material}.

\subsection{Clutter Reduction Module (CRM)}
The core of the proposed IRASNet lies in the novel feature-level clutter reduction mechanism (Fig.~\ref{fig:irasnet}, CRM), which simultaneously reflects the features of the target and shadow regions. Conventional clutter reduction methods reduce the intensity of clutter while preserving that of the target using the positional information of both the target and clutter regions \cite{li_HDANet, zhou2018sar, wang2019sar, choi2022fusion}. The methods aim to maximize SCR defined by:
\begin{align}
     SCR=20\log_{10}\left (\frac{\underset{p',q'\in I_T}{\text{max}}\,I_T(p',q')}{\frac{1}{N_C}\sum_{p''}\sum_{q''}I_C(p'',q'')}\right ),
\label{eq:imagescr}
\end{align}
where \(I_T(p',q') \in \mathbb{R}^{P_T\times Q_T}\) represents the target region. \(P_T\) and \(Q_T\) denote the width and height of the target region, respectively. Similarly, \(I_C(p'',q'') \in \mathbb{R}^{P_C \times Q_C}\) represents the clutter region. \(P_C\) and \(Q_C\) represent the width and height of the clutter region, respectively. Additionally, \(N_C\) denotes the total number of pixels in the clutter region. The maximization of Eq.~(\ref{eq:imagescr}) is implemented in \cite{choi2022fusion} using expressions:
\begin{align}
     I_T = I \otimes m_t ,
\label{eq:targetregion}
\end{align} where \(\otimes\) denotes element-wise multiplication and \( m_{t} \in \{0,1\}^{P\times Q} \) represent the masks for the target regions. Accordingly, the SCR is defined as \(\infty\). 

In the ATR study, information in the shadow region is determined and obtained as:
\begin{align}
     I_S = I \otimes m_s
\label{eq:shadowregion}
\end{align} to enhance class discrimination \cite{choi2022fusion}. Herein, \( I_S \in \mathbb{R}^{P \times Q} \) represents the shadow region, and \( m_{s} \in \{0,1\}^{P\times Q} \) denotes the mask of the shadow region. Therefore, to achieve higher ATR performance, maximizing SCR while preserving the information in the shadow region of the feature map is essential. Thus, to achieve the required results, the following optimization problem is solved.
\begin{equation}
    \begin{aligned}
        \underset{\theta_F}{\text{max}}& \quad 20\log_{10}\left(\frac{\underset{c,w,h\in F_{T}}{\max} F_{T}(c,w,h;\theta_F)}{\frac{1}{N_{F_C}}\sum\sum\sum F_{C}(c,w,h; \theta_F)}\right) \\
        &\quad \text{subject to} \quad \left\| F_{S}(\cdot;\theta_F) - H_{S} \right\|^2 \leq \epsilon ,
    \end{aligned}
\label{eq:crm_opt}
\end{equation}
where \(f(\cdot;\theta_F)\) denotes a CNN-based extractor with internal parameters \(\theta_F\) and \(N_{F_C}\) represents total number of pixels in the clutter feature. Additionally, \(F_T \in \mathbb{R}^{C \times W \times H}\), \(F_C \in \mathbb{R}^{C \times W \times H}\), and \(F_S \in \mathbb{R}^{C \times W \times H}\) represent the feature maps of the target, clutter, and shadow regions, respectively. Herein, Eq.~(\ref{eq:crm_opt}) represents the objectives that CRM aims to achieve. While maximizing SCR on the feature map, the shadow feature map must also minimize its difference from the encoded segmentation mask \(H_S \in \mathbb{R}^{C \times W \times H}\), which includes the positional information of the shadows. The method for obtaining \(H_S\) is introduced in the positional supervision task (PST). The CRM for solving Eq.~(\ref{eq:crm_opt}) comprises two sub-parts: \textit{1) positional supervision task (PST) and 2) attention task (AT)}.

Given an intermediate feature map \(F_{IN} \in \mathbb{R}^{C \times W \times H}\) as input, CRM infers a 2D target and shadow attention map \(Z_S \in \mathbb{R}^{1 \times W \times H}\). The overall feature-level clutter reduction process can be summarized as: 
\begin{align}
     F_{OUT} = Z_S \otimes F_{IN},
\label{eq:attention_mach}
\end{align}
where \(\otimes\) denotes the element-wise product. The attention map, which has high activation values for target and shadow, is multiplied by the input feature map. \(F_{OUT} \in \mathbb{R}^{C \times W \times H}\) represents the final refined output. Details of computing the attention map are discussed further. \\

\textit{1) Positional supervision task (PST)}\\
To enhance the SCR on the feature map, understanding the positional information of the target and shadow is crucial. In this section, we introduce a new approach called PST to focus on improving SCR by accurately reflecting the positional information of targets and shadows within feature maps, thereby enhancing spatial consistency. By introducing PST, we enable the dynamic integration of positional information of objects and shadows across domains.

PST is inspired by the key mechanism proposed in the pixel-level clutter reduction method introduced in Eq.~(\ref{eq:shadowregion}). The technique obtains target and shadow features by multiplying the mask features of the target and shadow with the input features.

One method to obtain mask features involves inputting target and shadow masks into the DL model. However, as shown in Fig.~\ref{fig:segerror_featureerror}, a loss can occur at the pixel level. Therefore, we adopt an approach where convolutional layers learn the masks by using target and shadow masks as GT. The masks are obtained using the method proposed in an existing study \cite{choi2022fusion}, where $m_t \in {\{0,1\}}^{P\times Q}$ and $m_s \in {\{0,1\}}^{P\times Q}$ are the target and shadow masks, respectively.

Initially, the input features are divided into three branches to incorporate positional information.
\begin{equation}
    \begin{aligned}
     f_e(F) &= ReLU(BN(f^{5\times5}(F)))\\
     F_{TM}&=f_e(f_e(f_e(F_{IN}))) \\
     F_{SM}&=f_e(f_e(f_e(F_{IN}))) \\
     F_{IN}'&= ReLU(BN(f^{1\times1}(F_{IN}))),
    \end{aligned}
\label{eq:3branch}
\end{equation}
where \(F_{TM}\), \(F_{IN}'\), and \( F_{SM} \in \mathbb{R}^{C\times W \times H}\) are the feature maps of the target mask, input, and shadow mask, respectively. Herein, \(f_e\) denotes the encoder function used to divide the input features into branches, \(BN\) denotes batch normalization, and \(f^{5\times5}\) represents a convolution operation with a filter size of \(5 \times 5\).

SAR images are significantly affected by speckle noise, which comprises high-frequency components. To reduce the impact of speckle noise and better capture positional information, we use \(5 \times 5\) convolution with a wide receptive field to obtain \(F_{TM}\) and \(F_{SM}\), which represent the target and shadow mask feature. Additionally, to minimize information loss from the input features and reduce complexity, we use \(1 \times 1\) convolution to obtain \(F_{IN}'\).

However, simply dividing the features into three branches is not sufficient to capture the information of the target and shadow. Therefore, we redefine the Mask GT encoding process to reflect the positional information of object and shadow regions. 

Previous studies \cite{zhang2020mask, tang2023salient, zhong2024mask} have integrated the positional information of objects into feature maps using Mask GT, primarily employing probability-based heatmaps and focal loss. In scenes containing multiple objects, probability-based heatmaps are used to roughly indicate object positions for tasks such as object detection and instance segmentation. Conversely, for classification tasks, which deal with images of a single object, the contour information of each element, such as objects, shadows, and noise, is crucial for accurate differentiation. However, using general probability-based heatmaps makes it difficult to accurately reflect the contour information.

Additionally, the CRM model integrates positional information into the feature maps \(F_{TM}\) and \(F_{SM}\). The Mask GT \(m_{t/s}\) exist in the space \( {\{0,1\}}^{P \times Q}\), while the feature maps \(F_{TM/SM}\) exist in the space \( \mathbb{R}^{C \times W  \times H}\), leading to modality mismatch. To resolve the discrepancy, we encode the Mask GT as:
\begin{equation}
    \begin{aligned}
      H_T = g(m_t; \theta_M)\\
      H_S = g(m_s; \theta_M),
    \end{aligned}
\label{eq:maskenc}
\end{equation}
where \(g(\cdot;\theta_M)\) represents the Mask GT encoder that shares the same structure as \(f(\cdot;\theta_F)\) to ensure the same modality with \(F_{TM}\) and \(F_{SM}\). However, it has different learnable parameters. The encoded Mask GTs \(H_T \in \mathbb{R}^{C \times W \times H}\) and \(H_S \in \mathbb{R}^{C \times W \times H}\) preserve the contour information of objects and shadows, matching the modality and scale of \(F_{TM}\) and \(F_{SM}\).

Subsequently, the loss function between the encoded GT and \(F_{TM}\) and \(F_{SM}\) is defined as:
\begin{equation}
    \begin{aligned}
      \mathcal{L}_{T} = -\frac{1}{N_S}\sum_{i=1}^{N_S}H_{T,i}\log(F_{TM,i})\\
      \mathcal{L}_{S} = -\frac{1}{N_S}\sum_{i=1}^{N_S}H_{S,i}\log(F_{SM,i}).
    \end{aligned}
\label{eq:maskloss}
\end{equation}
Accordingly, \(F_{TM}\) and \(F_{SM}\) in the CRM effectively integrate the positional information of targets and shadows.

Finally, \(F_{TM}\) and \(F_{SM}\) can be used to derive the target and shadow features through element-wise multiplication with \(F_{IN}'\):
\begin{equation}
    \begin{aligned}
      F_{T} = ReLU(F_{TM} \otimes F_{IN}')\\
      F_{S} = ReLU(F_{SM} \otimes F_{IN}'),
    \end{aligned}
\label{eq:target_shadow}
\end{equation}
where \(F_{T} \in \mathbb{R}^{C\times W \times H}\) and \(F_{S} \in \mathbb{R}^{C \times W \times H}\) represent the target and shadow region-dominant features, respectively. Additionally, the operation reflects the correlation with clutter regions in the input features, thereby maximizing the feature map SCR. Through the introduction of PST, effective handling of changes in key point positions that vary across domains is possible. Since the key point positions differ by domain \cite{zhang2023vsfa}, using masks is more effective than directly utilizing the segmented regions in the DG-ATR problem, as the approach helps to reduce domain-specific errors. \\

\textit{2) Attention task (AT)}\\
In an AT, an attention map is generated by utilizing key features from both the target and shadow regions, based on an attention mechanism \cite{woo2018cbam}. The attention map is designed to focus on areas within the target and shadow regions that contribute to the learning process while excluding clutter regions.
\begin{align}
     Z_S = Avgpool(&ReLU(f^{3\times3}([F_{T};F_{S}]))),
\label{eq:zs}
\end{align}
where \( Z_S \in \mathbb{R}^{1\times W\times H} \) represents the attention map, and \(Avgpool\) denotes global average pooling.

Finally, the clutter regions within the feature map are maximally reduced through PST, while preserving the shadow regions via Mask GT encoding. Through AT, the model is directed to focus on relevant areas within the target and shadow regions.

\subsection{Adversarial learning}
Adversarial learning, initially developed in generative adversarial networks (GAN) \cite{goodfellow2014generative}, has been utilized in DA \cite{dann, ganin2015unsupervised} and DG \cite{li2018deep}. IRASNet can be divided into a feature extractor \(f_F(\cdot;\theta_F)\) and a classifier \(f_Y(\cdot;\theta_Y)\). The model can be trained according to the classification loss as:
\begin{align}
    \mathcal{L}_{cls} = -\frac{1}{N}\sum_{i=1}^{N}\sum_{c=1}^{C}\mathbbm{1}_{[c=y_i]} \log(P(\hat{y_i}|x_i)),
\label{eq:class_loss}
\end{align}
where \(y_i\) is the class label and \(\hat{y_i}\) is the predicted value from the classifier. In addition to such components, adversarial learning introduces a domain discriminator \(f_D(\cdot;\theta_D)\), which is trained to discriminate the domains when the outputs of the feature extractor are inputted. Conversely, the feature extractor is trained to extract features that make it difficult for the domain discriminator to discriminate their domains. Thus, the extraction of domain-invariant features from multiple source domains is enabled, thereby generalizing the model for the unseen target domain. This is the core idea of DIRL using adversarial learning. The adversarial loss is defined as:
\begin{align}
    \mathcal{L}_{adv} = -\frac{1}{N}\sum_{i=1}^{N}\sum_{k=1}^{K}\mathbbm{1}_{[k=d_i]} \log(P(\hat{d_i}|x_i)),
\label{eq:adv_loss}
\end{align}
where \(K\) is defined as 2 (in this paper), \(d_i\) is the domain label, and \(\hat{d_i} \subset \mathcal{D}\) is the predicted value from the domain classifier. Both \(\mathcal{L}_{cls}\) and \(\mathcal{L}_{adv}\) are defined as negative log-likelihood loss functions.

During training, weight updates for the feature extractor, domain discriminator, and label classifier are performed differently, expressed as:
\begin{equation}
    \begin{aligned}
        &\theta_F \gets \theta_F - \mu \left( \frac{\partial \mathcal{L}_{cls}}{\partial \theta_F} - \lambda \frac{\partial \mathcal{L}_{adv}}{\partial \theta_F} \right) \\ 
        &\theta_Y \gets \theta_Y - \mu \frac{\partial \mathcal{L}_{cls}}{\partial \theta_Y} \\ 
        &\theta_D \gets \theta_D - \mu \frac{\partial \mathcal{L}_{adv}}{\partial \theta_D}.
    \end{aligned}
\label{eq:weight_update}
\end{equation}

To introduce negative weights into the learning process, we use the gradient reversal layer (GRL) \cite{ganin2015unsupervised}. The GRL does not have any parameters except for a hyper-parameter \(\lambda\) (which is not updated via backpropagation). During the forward pass, the GRL acts as an identity transformation. However, during the backward pass, the GRL multiplies the gradient received from the subsequent layer by \(-\lambda\) and passes it to the previous layer. IRASNet achieves enhanced generalization performance by simultaneously learning domain-invariant and clutter-reduced features through the combination of CRM and adversarial learning.

\begin{table*}[t]
\centering
\caption{Training and test SAR samples under four scenarios of experimental setup}
\label{tab:scenario}
\renewcommand{\arraystretch}{1.1} 
\resizebox{0.75\textwidth}{!}{
\begin{tabular}{c|cc|cc|cc|cc}
\toprule
\toprule
                  & \multicolumn{2}{c|}{\textbf{Scenario 1}} & \multicolumn{2}{c|}{\textbf{Scenario 2}} & \multicolumn{2}{c|}{\textbf{Scenario 3}}        & \multicolumn{2}{c}{\textbf{Scenario 4}} \\ \cline{2-9} 
\multirow{2}{*}{} & \textbf{Train}           & \textbf{Test}          & \textbf{Train}          & \textbf{Test}           & \textbf{Train}        & \textbf{Test}                    & \textbf{Train}        & \textbf{Test}            \\
                  & Synthetic       & Measured      & Synthetic      & Measured       & Synthetic    & \begin{tabular}[c]{@{}c@{}}SCR Fluct.\\ (-3 $-$ +3)\end{tabular} & Synthetic    & \begin{tabular}[c]{@{}c@{}}Unknown\\ Clutter\end{tabular} \\ \midrule
Dep. Angle                  & 14$-$16      & 17            & 14$-$17     & 14$-$17     & 14$-$17   & 14$-$17              & 14$-$17   & 14$-$17      \\ \midrule
2S1               & 116    & 58   & 174   & 174   & 174 & 174          & 174 & 174    \\
BMP2              & 55              & 52            & 107            & 107            & 107          & 107                     & 107          & 107             \\
BTR70             & 43              & 49            & 92             & 92             & 92           & 92                      & 92           & 92              \\
M1                & 78              & 51            & 129            & 129            & 129          & 129                     & 129          & 129             \\
M2                & 75              & 53            & 128            & 128            & 128          & 128                     & 128          & 128             \\
M35               & 76              & 53            & 129            & 129            & 129          & 129                     & 129          & 129             \\
M548              & 75              & 53            & 128            & 128            & 128          & 128                     & 128          & 128             \\
M60               & 116             & 60            & 176            & 176            & 176          & 176                     & 176          & 176             \\
T72               & 56              & 52            & 108            & 108            & 108          & 108                     & 108          & 108             \\
ZSU23             & 116             & 58            & 174            & 174            & 174          & 174                     & 174          & 174             \\ \midrule
Total             & 806             & 539           & 1345           & 1345           & 1345         & 1345                    & 1345         & 1345            \\ \bottomrule
\end{tabular}%
}
\end{table*}

\section{Experiments}
\subsection{Dataset Description}
\begin{figure}[t]
\centering
\includegraphics[width=\linewidth]{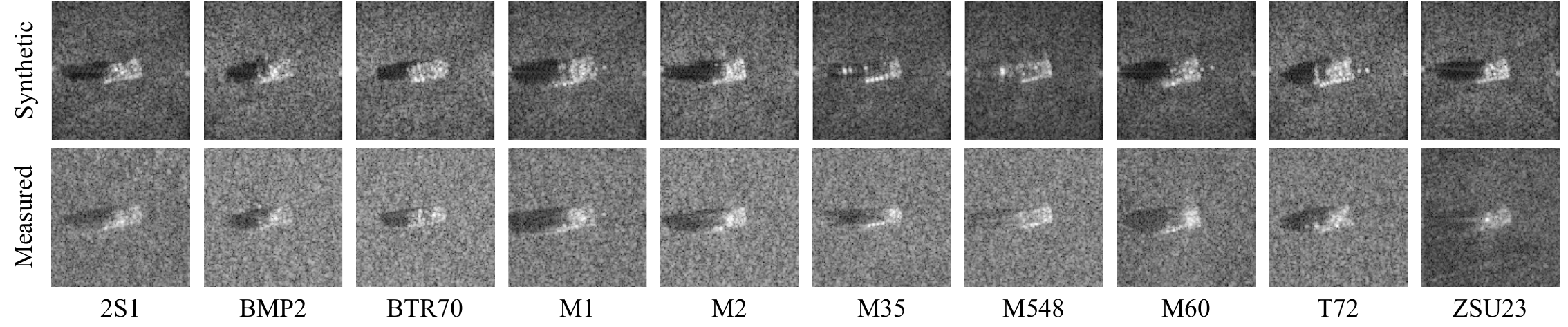}
\caption{Paired synthetic and measured images in the SAMPLE dataset.}
\vspace{-1.2em}
\label{fig:smpl_example}
\end{figure}

\textit{1) SAMPLE dataset:} To verify the DG performance of the proposed IRASNet, we adopted the SAMPLE dataset \cite{smpl} that included both synthetic and measured data. In the dataset, synthetic data were generated using sophisticated CAD models and asymmetric ray tracing techniques \cite{smpl}, while the measured data were directly obtained from the MSTAR dataset \cite{zhang2023vsfa}. The SAMPLE dataset comprised 10 ground targets captured under various SOC conditions, which were categorized into 10 different ground vehicle target classes: 2S1, BMP2, BTR70, M1, M2, M35, M548, M60, T72, and ZSU23. The paired images of synthetic and measured data for each class are shown in Fig.~\ref{fig:smpl_example}. The SAR target images were obtained using an X-band HH-polarized SAR with a resolution of 0.3×0.3 m, and image size of 128×128 pixels. The depression angle ranged from 14° to 17°, and the azimuth angle ranged from 10° to 80°. The entire SAMPLE dataset contained 1,345 paired SAR synthetic and measured image pairs with the same imaging parameters, including depression and azimuth angles. \\

\textit{2) SCR fluctuation:} In SAR-ATR, performance degraded when clutter overwhelmed the target signal, making performance evaluation under varying SCR conditions crucial. Additionally, the core concept of the proposed IRASNet involves clutter reduction at the feature level. Therefore, to assess the clutter reduction performance of IRASNet and compare it with various feature-level clutter reduction methods, we created a new test set by adjusting the SCR of the measured data in the SAMPLE dataset. First, we obtained \(I_T\) and \(I_C\) from the measured SAR image \(x^{mea}\) using the target/shadow segmentation algorithm proposed in a previous study \cite{choi2022fusion}. 

Further, we calculated SCR using Eq.~\ref{eq:imagescr} and adjusted the intensity of the clutter region according to the variation value to generate a new \(x^{scr}\). The dataset comprised 13 test sets, with SCR increasing from $-$3 dB to 3 dB in increments of 0.5 dB. The newly generated SAR images with varying SCR are presented in \textit{Supplementary Material}. Notably, SCR variations were only applied to the measured data to evaluate models trained on synthetic data.\\

\textit{3) Different background clutter:} The MSTAR chip and SAMPLE datasets, currently used as benchmark datasets, included only clean ground clutter among various types of clutter. However, the target data required for ATR in real-world applications included clean ground clutter along with complex ground clutter caused by trees, grass, and the shadows they form \cite{chen2016target}. To more accurately evaluate ATR performance in actual operational environments, providing datasets that include complex types of clutter is essential. Therefore, we utilized the MSTAR public clutter data including both rural and urban clutter. Similarly, by segmenting the target/shadow in the measured data, we synthesized and cropped the images with the MSTAR public clutter data to obtain 128$\times$128 images. The data generated through this method is referred to as unknown clutter. The newly generated SAR images with unknown clutter are presented in \textit{Supplementary Material}. Along with SCR fluctuation, an unknown background clutter was applied only to the measured data to evaluate models trained on synthetic data. The dataset allowed us to assess clutter reduction performance in relatively complex backgrounds and evaluate it from the perspective of DG.\\

\textit{4) Experimental scenario:} Four main experimental scenarios were considered. In scenario 1, training data comprised synthetic data with a depression angle ranging from 14° to 16°, while test data comprised measured data with a depression angle of 17°. The scenario was closer to real-world conditions as there was no one-to-one correspondence between the synthetic and measured data. In scenario 2, the training data included synthetic data with depression angles ranging from 14° to 17°, and the test data used measured data with the same range of depression angles. The two scenarios allowed us to evaluate the performance of various models from the perspective of DG.

The remaining two scenarios were designed to evaluate the potential of feature-level clutter reduction. In scenario 3, the synthetic data used in scenario 2 was utilized as training data, and the test data comprised 13 measured datasets with SCR fluctuation increasing from $-$3 dB to 3 dB in increments of 0.5 dB. Since the training data were fixed at an SCR fluctuation of 0 dB, the varied SCR in the test data made the recognition task more challenging. Lastly, scenario 4 was designed to evaluate clutter reduction performance in complex backgrounds. The same synthetic data from scenario 2 was used as training data, and the test data comprised measured datasets with complex backgrounds. As the DL model was trained only on ground clutter, the scenario assessed its ability to reduce clutter in more complex backgrounds. Detailed information on these four scenarios is provided in Table~\ref{tab:scenario}. The augmented dataset was consistently applied to all comparison models across the experiments.

\begin{table}[t]
\centering
\caption{Experimental results under scenario 1 and 2\\according to generalization methods}
\label{tab:comparion1}
\renewcommand{\arraystretch}{1.4} 
\resizebox{0.95\columnwidth}{!}{%
\begin{tabular}{ccccc}
\toprule
\toprule
\multirow{2}{*}{} & \multirow{2}{*}{\textbf{Speciality}} & \multirow{2}{*}{\textbf{Model}} & \multicolumn{2}{c}{\textbf{Accuracy {[}\%{]}}} \\ \cline{4-5} 
                  &                                     &                                 & \textbf{\begin{tabular}[c]{@{}c@{}}Under\\ Scenario 1\end{tabular}} & \textbf{\begin{tabular}[c]{@{}c@{}}Under\\ Scenario 2\end{tabular}} \\ \midrule
\multirow{11}{*}{\begin{tabular}[c]{@{}c@{}}Pixel\\level\\ Gen.\end{tabular}} 
                  & \multirow{6}{*}{\begin{tabular}[c]{@{}c@{}}Vision\\ algorithm\end{tabular}} & EfficientNet \cite{tan2019efficientnet}  & 94.99\% & 92.04\% \\
                  &                                     & MobileNet V2 \cite{sandler2018mobilenetv2} & 90.35\% & 90.63\% \\
                  &                                     & MobileNet V3 \cite{howard2019searching} & 91.09\% & 91.38\% \\
                  &                                     & ResNet18 \cite{resnet}      & 96.29\% & 91.30\% \\
                  &                                     & ResNet50 \cite{resnet}     & 92.57\% & \underline{93.76\%} \\
                  &                                     & ResNext \cite{xie2017aggregated}      & 93.69\% & 93.75\% \\ \cline{2-5} 
                  & \multirow{4}{*}{\begin{tabular}[c]{@{}c@{}}SAR-ATR\\ algorithm\end{tabular}} & AconvNet  \cite{chen2016target}    & 89.42\% & 87.14\% \\
                  &                                     & AM-CNN \cite{amcnn}       & 94.62\% & 93.53\% \\
                  &                                     & ASIR-Net \cite{yu2021lightweight}     & 92.21\% & 93.75\% \\
                  &                                     & TAI-SARNet \cite{ying2020tai}    & 85.52\% & 83.49\% \\ \cline{2-5} 
                  & DG-ATR & Ensemble \cite{inkawhich2021bridging}& \underline{95.06\%} & - \\ \midrule
\multirow{8}{*}{\begin{tabular}[c]{@{}c@{}}Feature\\level\\ Gen.\end{tabular}} 
                  & \multirow{7}{*}{\begin{tabular}[c]{@{}c@{}}Vision \\ algorithm\\ for DG\end{tabular}} & DANN \cite{dann}         & 91.65\% & 91.45\% \\
                  &                                     & MMLD  \cite{matsuura2020domain}        & 91.09\% & 91.67\% \\
                  &                                     & DANN+DFF  \cite{lin2023deep}    & 92.39\% & 93.01\% \\
                  &                                     & MMD-AAE  \cite{Li_2018_CVPR}     & 90.35\% & 91.97\% \\
                  &                                     & CAADA  \cite{rahman2020correlation}       & 89.42\% & 90.63\% \\
                  &                                     & Deep CORAL$^{1}$ \cite{sun2016deep} & 94.99\% & 92.93\% \\
                  &                                     & Deep CORAL$^{2}$ \cite{sun2016deep} & 93.69\% & 92.19\% \\ \cline{2-5} 
                  & DG-ATR & IRASNet \textbf{(Proposed)} & \textbf{97.40\%} & \textbf{96.72\%} \\ \bottomrule
\end{tabular}%
}
\vspace{-1.2em}
\end{table}

\subsection{Comparison experiments}
First, using scenarios 1 and 2, we conducted comparative experiments from a DG perspective. We comprehensively investigated DG-ATR performance based on various network architectures, including backbone models specialized for SAR images \cite{chen2016target, amcnn, yu2021lightweight, ying2020tai} and basic SAR-related models developed in the field of image classification \cite{tan2019efficientnet, sandler2018mobilenetv2, howard2019searching, resnet, xie2017aggregated}. The experiment was performed to verify whether the domain gap reduced at the pixel level through data augmentation was overcome at the feature level. 

To thoroughly validate the effectiveness of our method, we compared IRASNet not only with specialized DG frameworks for SAR images \cite{inkawhich2021bridging} but also with state-of-the-art DG algorithms designed for optical images \cite{matsuura2020domain, lin2023deep, Li_2018_CVPR, rahman2020correlation, sun2016deep}. Our study was conducted by training exclusively on 100$\%$ synthetic data and testing on measured data. Therefore, algorithms that included measured data for training were excluded from our experiments as they were beyond the scope of this study. Due to the unavailability of source code for the algorithm proposed in a previous study \cite{inkawhich2021bridging}, the results for scenario 1 were directly quoted from the original paper, while the results for scenario 2 were excluded as the experiment was not conducted in that study. The performances of the remaining algorithms were implemented and evaluated based on the available source codes, and the results for the two experimental scenarios are presented in Tables~\ref{tab:comparion1} and ~\ref{tab:comparion2}, respectively. The results were reported as the average of five independent experiments to ensure statistical reliability, with the best performance highlighted in \textbf{bold} and the second-best performance \underline{underlined} in the tables.

\begin{table}[t]
\centering
\caption{Experimental results under scenario 1 and 2\\according to clutter mitigation methods}
\label{tab:comparion2}
\renewcommand{\arraystretch}{1.2} 
\resizebox{0.85\columnwidth}{!}{
\begin{tabular}{ccccc}
\toprule
\toprule
\textbf{Speciality} & \textbf{Model} & \multicolumn{2}{c}{\textbf{Accuracy [\%]}} \\
\cmidrule(lr){3-4}
& & \textbf{\begin{tabular}[c]{@{}c@{}}Under\\ Scenario 1\end{tabular}} & \textbf{\begin{tabular}[c]{@{}c@{}}Under\\ Scenario 2\end{tabular}} \\ \midrule
\multirow{3}{*}{\begin{tabular}[c]{@{}c@{}}Pixel Level\\ CR\end{tabular}}                
& SAR-IFTS  \cite{choi2022fusion}                                                   & 80.33\%                                                             & 80.07\%                                                             \\
                                     & ESENet \cite{wang2019sar}                                                          & 74.58\%                                                             & 75.76\%                                                             \\
                                     & LM-BN-CNN \cite{zhou2018sar}                                                   & 71.80\%                                                             & 75.39\%                                                             \\ \midrule
Clutter Robust                                  
& CFA \cite{cfa}                                                         & 86.64\%                                                             & 86.17\%                                                             \\ \midrule
\multirow{2}{*}{\begin{tabular}[c]{@{}c@{}}Feature Level\\ CR \end{tabular}}                
& HDANet \cite{li_HDANet}                                                      & \underline{92.21\%}                                                             & \underline{90.92\%}                                                             \\
                                     & IRASNet \textbf{(Proposed)} & \textbf{97.40\%}                                                    & \textbf{96.72\%}                                                    \\ \bottomrule
\end{tabular}%
}
\end{table}

As shown in Table~\ref{tab:comparion1}, despite reducing the domain gap at both the pixel and feature levels, a comparison of ATR results from existing methods indicated that only algorithms using skip connections, such as \cite{resnet} and \cite{xie2017aggregated}, and those applying attention mechanisms, such as \cite{amcnn}, achieved satisfactory performance. Skip connections and attention mechanisms enhanced the activation of targets and enabled focused computations by adding operations to existing features. However, such methods did not completely overcome the domain gap.

\begin{figure*}[t]
\centering
\includegraphics[width=\linewidth]{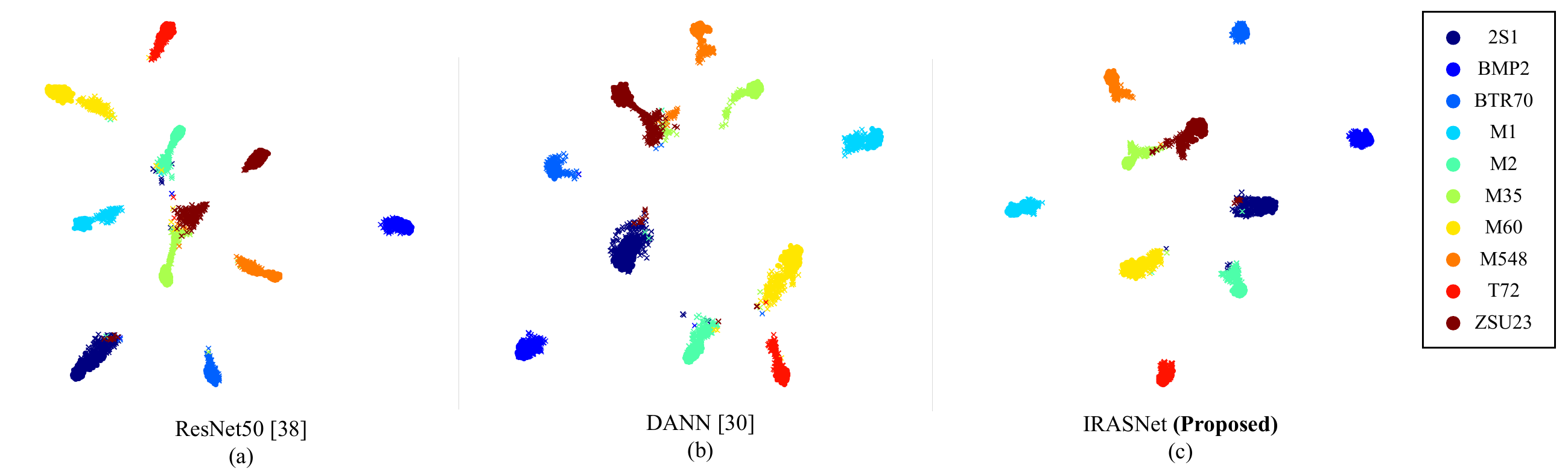}
\caption{Visualization of feature distribution using t-SNE embeddings. (a) ResNet50 \cite{resnet}. (b) DANN \cite{dann}. (c) IRASNet. Features extracted from samples of the same category are shown in the same color, with synthetic (represented by o) and measured (represented by x) images.}
\vspace{-0.7em}
\label{fig:tsne}
\end{figure*}

As shown in Fig.~\ref{fig:tsne}, we visualized the feature space using t-SNE \cite{van2008visualizing} for ResNet \cite{resnet}, which recorded the highest performance, and DANN \cite{dann}, the baseline for IRASNet. We observed that in the latent spaces of ResNet \cite{resnet} and DANN \cite{dann}, the alignment of four classes was not well-matched, and the distinction between classes was visually lacking. Thus, the domain gap had not been fully overcome in the feature space. Contrarily, the proposed IRASNet successfully reduced the domain gap through feature-level clutter reduction, allowing for both domain alignment and discrimination in the t-SNE visualization. Moreover, IRASNet achieved state-of-the-art performance, being 2.34$\%$ higher than the existing technique \cite{inkawhich2021bridging} under scenario 1 and 2.97$\%$ higher than the existing models under scenario 2. The results support our motivation to reduce clutter at the feature level for DG-ATR.

Notably, this study focuses on analyzing the impact of clutter in the pattern analysis of radar images. Compared to the baseline DANN \cite{dann}, the results demonstrated that differences in distribution at the feature level still negatively affected discrimination due to the influence of clutter. Furthermore, the influence of feature-level clutter reduction significantly enhanced discrimination in the latent space and suggested the possibility of overcoming the domain alignment problem.

Additionally, using scenarios 1 and 2, we conducted comparative experiments from the perspective of clutter reduction. We compared IRASNet with algorithms such as those presented by previous studies \cite{choi2022fusion, wang2019sar},\cite{zhou2018sar}, which processed by segmenting target and shadow areas in the preprocessing stage. Further, algorithm comparison was performed with that of previous studies \cite{li_HDANet} and \cite{cfa}, which employed feature-level clutter reduction methods and clutter-robust learning methods, respectively. As shown in Table~\ref{tab:comparion2}, feature-level clutter reduction methods generally achieved higher performance compared to pixel-level clutter reduction methods. Despite the mitigated distribution differences at the pixel level through data augmentation, the result demonstrated, through various algorithms, that the performance was significantly affected by the information lost at the input stage, as shown in Fig.~\ref{fig:segerror_featureerror}.

Furthermore, the proposed IRASNet showed an approximate improvement of 6$\%$ under both scenarios 1 and 2 compared to existing feature-level clutter reduction methods and clutter-robust learning methods. The result supported our motivation for the need for domain-invariant feature learning in addition to clutter reduction. The significantly improved ATR performance in the two experimental scenarios fully demonstrated the effectiveness of our algorithm.

\subsection{Feature-level clutter reduction performance comparison}
This section describes experiments conducted using scenarios 3 and 4 to evaluate the performance of feature-level clutter reduction in various background clutter situations. First, by introducing SCR fluctuations into the SAMPLE measured dataset, we assessed the robustness of the algorithm under extreme SCR conditions by evaluating ATR performance and clutter reduction performance in various SCR scenarios. Additionally, we analyzed ATR performance and clutter reduction in situations with background clutter patterns not included in the training, such as grass, trees, and shadows, to evaluate the generalization ability of atoms to unexpected background clutter.

Finally, we applied explainable AI techniques, specifically SHAP \cite{lundberg2017unified}, to determine which parts of the input image contributed to the DL model under SCR fluctuations and various background clutter conditions. The analysis assisted us in understanding the performance results from the previous two experiments. To evaluate the performance of the proposed CRM, we comprehensively investigated its performance alongside existing feature level clutter reduction methods, such as \cite{li_HDANet}, and methods for clutter-robust SAR-ATR at the feature level \cite{cfa}. As described in Table~\ref{tab:scenario}, all algorithms used in the experiments were trained on the SAMPLE synthetic dataset and augmented data without SCR fluctuations. \\

\begin{figure*}[t]
\centering
\includegraphics[width=\linewidth]{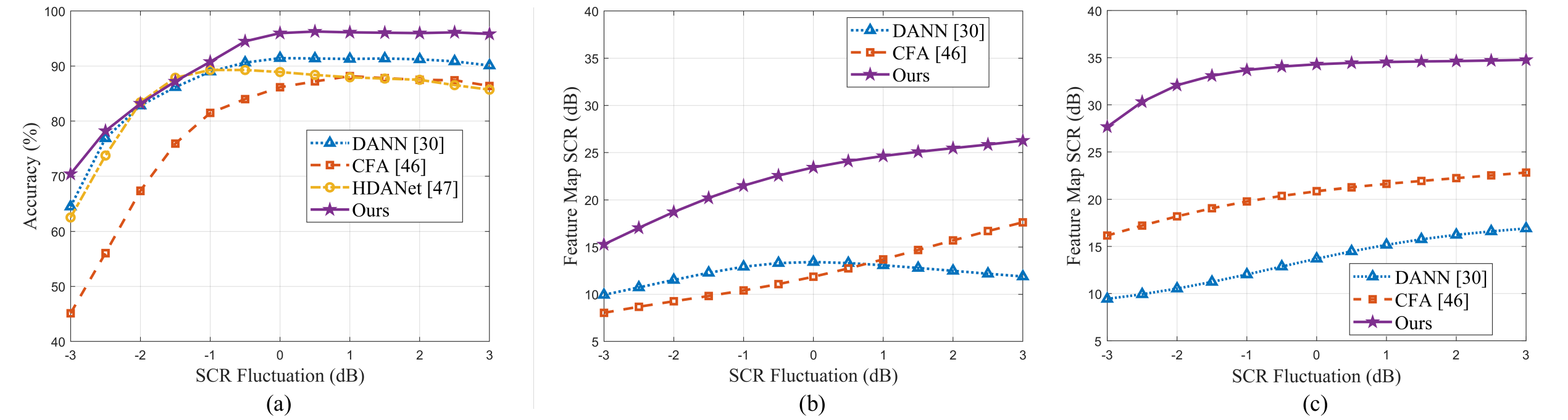}
\caption{(a) Accuracy on the SCR fluctuation test set under scenario 3. (b) Feature map SCR in layer 1 on the SCR fluctuation test set under scenario 3. (c) Feature map SCR in layer 2 on the SCR fluctuation test set under scenario 3.}
\vspace{-0.8em}
\label{fig:scr_flctuation_results}
\end{figure*}

\textit{1) SCR fluctuation dataset:} As shown in Fig.~\ref{fig:scr_flctuation_results}, performance decreased when the SCR was low and improved as the SCR increased. The proposed IRASNet minimized performance degradation even in low SCR conditions and enhanced performance in high SCR conditions, achieving the best performance in most test sets with SCR fluctuations. Notably, even with the same training dataset, IRASNet improved performance by 7.88$\%$ compared to an existing method \cite{li_HDANet} in a $-$3 dB dataset, where the brightness difference between the target and clutter was minimal. Thus, a further improvement in feature-level clutter reduction performance over existing methods is required. Additionally, compared to an existing study \cite{cfa}, which used clutter-robust learning methods, IRASNet showed a 25.28$\%$ improvement in performance. Thus, the approach of removing clutter was more effective in reducing the influence of clutter and achieving high performance than learning clutter. Additionally, the proposed IRASNet achieved over 5$\%$ better performance than all other methods, even in scenarios where SCR was higher than in the training dataset.

As presented in Table~\ref{tab:scr_fluct_result}, a slight difference between SCR fluctuations at 0 and 3 dB was observed. A previous study \cite{cfa} recorded 0.22$\%$ higher performance at 3 dB than at 0 dB. Thus, in high SCR situations, the clutter-robust learning approach can reduce the impact of clutter, thereby improving performance. However, \cite{cfa} showed significant performance variations in low SCR conditions and low performance in high SCR conditions, as it failed to effectively extract the intrinsic features of the target.

Conversely, another study \cite{li_HDANet} used feature-level clutter reduction and recorded 3.2$\%$ lower performance at 3 dB compared to that at 0 dB. The result suggested that the previous study \cite{li_HDANet} failed to perform proper clutter reduction in high SCR situations and relied on clutter for decision-making. Contrarily, the proposed IRASNet reduced the difference to 0.15$\%$, improving clutter reduction performance compared to existing methods and largely resolving the clutter learning problem. Additionally, IRASNet better extracted the intrinsic features of the target, enhancing discrimination and achieving high performance.

Furthermore, the SCR-fluctuated test set caused another domain gap. DANN\cite{dann} only used the DG technique by employing DIRL and achieved high performance above 85$\%$ at $-$1.5 dB and above, except in very low SCR conditions. Contrarily, the proposed IRASNet suggested that the additional incorporation of clutter reduction further enhanced the generalization performance of ATR.

The proposed IRASNet was designed to maximize SCR in the feature map, as expressed in Eq.~\ref{eq:crm_opt}. For numerical validation, we examined the SCR of the feature maps under SCR fluctuation. HDANet \cite{li_HDANet} extracted features with omitted details due to the deep encoder layers, making it impossible to identify the exact locations of targets, shadows, and clutter. Thus, the SCR of the feature map could not be calculated and was excluded from the analysis. As shown in Fig.~\ref{fig:scr_flctuation_results}, the proposed IRASNet improved the feature map SCR by approximately 10 dB in layer 1 and over 19 dB in layer 2 compared to the conventional DANN \cite{dann}. Thus, the optimization problem in Eq.~\ref{eq:crm_opt} has been successfully achieved.

\begin{table}[t]
\centering
\caption{Average accuracy and accuracy difference between 0 and $-$3 dB and between 0 and 3 dB datasets obtained on the SCR fluctuation test set under scenario 3.}
\label{tab:scr_fluct_result}
\renewcommand{\arraystretch}{1.7} 
\resizebox{0.9\columnwidth}{!}{%
\begin{tabular}{cccccc}
\toprule
\toprule
\multicolumn{2}{c}{\textbf{Testset}} & \textbf{DANN \cite{dann}} & \textbf{CFA \cite{cfa}}     & \textbf{HDANet \cite{li_HDANet}} & \textbf{IRASNet} \\ 
\midrule
\multirow{4}{*}{\begin{tabular}[c]{@{}c@{}}\textbf{SCR}\\ \textbf{Fluct.}\end{tabular}} & -3dB             & \underline{64.51\%}      & 45.13\%          & 62.53\%          & \textbf{70.41\%} \\
                                     & 0dB              & \underline{91.45\%}      & 86.17\%          & 88.92\%          & \textbf{95.99\%} \\
                                     & 3dB              & \underline{90.11\%}      & 86.39\%          & 85.72\%          & \textbf{95.84\%} \\
                                     & \textbf{Average} & \underline{86.75\%}    & 78.51\%          & 84.54\%          & \textbf{90.52\%} \\ 
\midrule
\multirow{2}{*}{\textbf{Diff.}} & $\Delta$ $Acc_{(0\mathrm{dB},-3\mathrm{dB})} (\downarrow) $ & 26.94\% & 41.04\% & \underline{26.39\%}   & \textbf{25.58\%} \\
                                     & $\Delta$ $Acc_{(0\mathrm{dB},3\mathrm{dB})} (\downarrow)$& 1.34\%  & \textbf{-0.22\%} & 3.20\%          & \underline{0.15\%}      \\ 
\bottomrule
\end{tabular}%
}
\vspace{-1.2em}
\end{table}
CFA \cite{cfa} employed robust learning rather than reducing clutter at the feature level through contrastive learning. Therefore, even with a low feature map SCR, the influence of clutter was reduced and consistent results were obtained even when the SCR fluctuation was +3dB, as shown in Fig.~\ref{fig:scr_flctuation_results}. However, in cases not present in the training data (such as $-$3dB), the performance dropped significantly, failing to guarantee consistent results. Therefore, a strong dependency on the training data was observed. \\

\textit{2) Unknown clutter dataset:} We analyzed ATR and clutter reduction performances in situations with various patterns of background clutter, such as grass, trees, and shadows, that were not included in the training data to evaluate the generalization ability of the algorithm to unexpected background clutter. As presented in Table~\ref{tab:unknown_results}, the feature-level clutter reduction methods, IRASNet and HDANet \cite{li_HDANet}, demonstrated the best performance. While DANN \cite{dann} showed the second-highest performance when the background clutter matched the training conditions, HDANet \cite{li_HDANet} outperformed DANN \cite{dann} in other scenarios. Thus, conventional ATR algorithms were vulnerable in real-world operating environments where clutter not included in the training data was present. Additionally, IRASNet improved the performance of HDANet \cite{li_HDANet} by over 6$\%$, proving its significantly enhanced adaptability to various background clutter in actual operating environments.

Both clutter-invariant learning and feature-level clutter reduction methods showed performance degradation when new clutter information, not used in training, was introduced because the introduction of new clutter information caused another distribution shift. Since CNNs extracted features by considering the correlation between targets and clutter, even when clutter reduction or invariant learning was performed, it affected the decision boundary, leading to performance degradation. However, the proposed IRASNet maintained high discriminability even when clutter changed, by better extracting the intrinsic features of the target through mask encoding, resulting in over a 6$\%$ performance improvement compared to existing methods.

\begin{table}[]
\centering
\caption{Experimental results under scenario 2 and 4}
\label{tab:unknown_results}
\renewcommand{\arraystretch}{1.6} 
\resizebox{0.9\columnwidth}{!}{%
\begin{tabular}{ccccc}
\toprule
\toprule
                                                          & \textbf{DANN \cite{dann}} & \textbf{CFA \cite{cfa}} & \textbf{HDANet \cite{li_HDANet}} & \textbf{IRASNet} \\ 
\midrule
\textbf{SMPL Measured}   & \underline{91.45\%} & 86.17\%      & 88.92\%         & \textbf{95.99\%} \\
\textbf{Unknown Clutter} & 86.47\%       & 84.68\%      & \underline{87.73\%}   & \textbf{93.38\%} \\ 
\bottomrule
\end{tabular}%
}
\vspace{-1.2em}
\end{table}

\begin{figure*}[t]
\centering
\includegraphics[width=\linewidth]{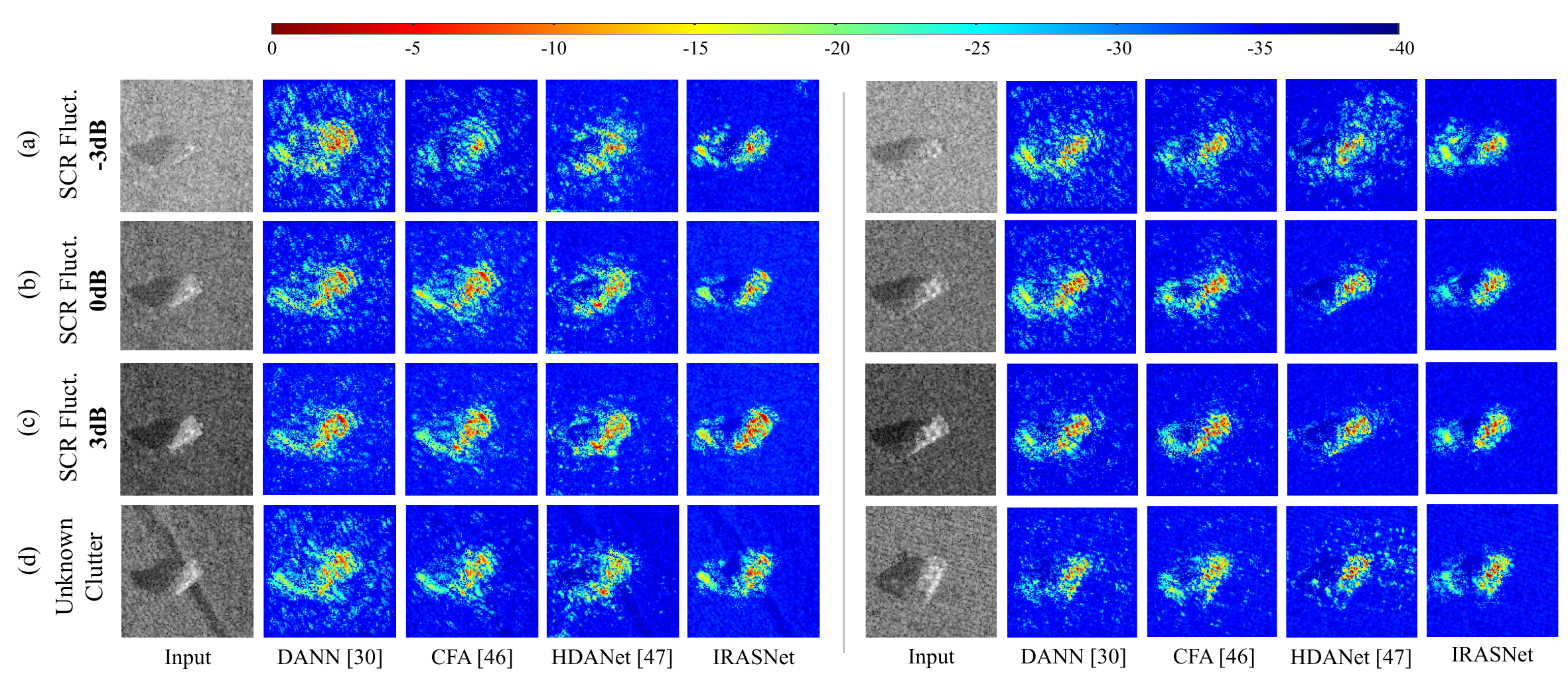}
\caption{Visualization of input image contributions in DL models using SHAP. (a) Test set of SCR Fluctuation -3dB. (b) Test set of SCR Fluctuation 0dB (SAMPLE Measured dataset). (c) Test set of SCR Fluctuation 3dB. (d) Test set of Unknown Clutter}
\vspace{-1em}
\label{fig:shap_final}
\end{figure*}

\subsection{XAI analysis}
To analyze the impact of clutter on the decisions made by DL models, we introduced the explainable AI technique known as SHAP \cite{lundberg2017unified}. SHAP is a method based on Shapley values from game theory to calculate the contribution of each feature to a prediction of the model, helping to interpret the model output \cite{lundberg2017unified}. In SHAP, a larger absolute SHAP value indicated that a particular feature had a greater influence on the decision of the model. Therefore, we used the absolute values of SHAP to identify the features that played important roles in decision-making for the target, shadow, and clutter areas. Additionally, to capture areas with very small SHAP values, we applied a dB scale and visualized the range from 0 to $-$40 dB. Values smaller than $-$40dB were considered to have negligible influence on the prediction of the model and were excluded from the analysis.

To evaluate clutter reduction performance, we analyzed the SAMPLE measured data using SHAP values. Table \ref{tab:shap} shows the results of calculating the sum of SHAP values for each area across all sample measured data and then computing their ratios. Higher SHAP values for the target and shadow areas indicate better performance, while lower SHAP values in the clutter areas suggest better clutter reduction. Across all datasets, it was observed that the proposed IRASNet reduced the influence of clutter and focused more on target and shadow areas compared to all existing methods.

To visually confirm the results, Fig.~\ref{fig:shap_final} shows the SHAP value visualizations for each pixel in images corresponding to the m60 and m1 classes, under the same target conditions for SAMPLE measured, SCR fluctuation $-$3 dB, +3 dB, and different background clutter. According to Table~\ref{tab:shap} and Fig.~\ref{fig:shap_final}, DANN \cite{dann} did not perform clutter reduction, resulting in contributions across all areas of the image, not just the target region. Additionally, CFA \cite{cfa}, through clutter-robust learning, focused more on target and shadow areas compared to DANN \cite{dann}. However, since the values of the clutter area remained in the feature space, it still relied on the clutter area for decision-making. While HDANet \cite{li_HDANet} significantly reduced the influence through feature-level clutter reduction, it still made decisions using clutter areas.

Contrarily, the proposed IRASNet successfully performed feature-level clutter reduction across four different clutter scenarios, significantly reducing the contribution of clutter compared to other methods. Thus, the proposed method relied minimally on clutter for decision-making and instead based its decisions on target and shadow computations. IRASNet was trained using mask information containing shape details for each class during mask encoding, which allowed it to effectively reduce new clutter not included in the class. Therefore, mask encoding successfully reflected the shape and positional information of the target and shadow. According to Table.~\ref{tab:shap}, in situations where various background clutter patterns such as grass, trees, and shadows were not included in the training, pixel-level clutter reduction caused a loss of information on targets and shadows, resulting in performance degradation. However, we addressed the issue by incorporating CRM.

Moreover, as seen in Fig. \ref{fig:shap_final}, the models primarily focused on the left part of the shadow when making decisions because, in the SAMPLE dataset, the difference in brightness between the shadow and clutter made the shadow more prominent. The characteristic acted as an important feature in CNNs, leading most networks to make decisions based on the left part of the shadow. The proposed CRM was trained considering the shadow mask through \(F_{SM}\), enhancing discriminability and positively affecting performance. While changes in the brightness of clutter and shadow affected the decisions of all networks, the emphasis on shadows positively influenced the results even when some clutter was considered. SHAP analysis revealed a tendency to closely examine the ends of shadow areas, visually demonstrating that shadows contained important information, as mentioned in \cite{choi2022fusion}. The information played a crucial role in enhancing class-specific discriminability.

\begin{table}[t]
\centering
\caption{Proportion of the sum of SHAP values \\for each area in the SAMPLE measured dataset.}
\label{tab:shap}
\renewcommand{\arraystretch}{1.5} 
\resizebox{0.9\columnwidth}{!}{%
\begin{tabular}{ccccc}
\toprule
\toprule
\textbf{}                         & \hspace{0.1cm}\textbf{DANN \cite{dann}}\hspace{0.1cm} & \hspace{0.1cm}\textbf{CFA \cite{cfa}}\hspace{0.1cm} & \hspace{0.1cm}\textbf{HDANet \cite{li_HDANet}}\hspace{0.1cm} & \hspace{0.1cm}\textbf{IRASNet}\hspace{0.1cm} \\ 
\midrule
\textbf{Target} (\(\uparrow\))    & \hspace{0.1cm}0.37\hspace{0.1cm}          & \hspace{0.1cm}0.41\hspace{0.1cm}         & \hspace{0.1cm}\underline{0.48}\hspace{0.1cm}            & \hspace{0.1cm}\textbf{0.57}\hspace{0.1cm}    \\
\textbf{Shadow} (\(\uparrow\))    & \hspace{0.1cm}0.27\hspace{0.1cm}          & \hspace{0.1cm}0.31\hspace{0.1cm}         & \hspace{0.1cm}\underline{0.33}\hspace{0.1cm}            & \hspace{0.1cm}\textbf{0.38}\hspace{0.1cm}    \\
\textbf{Clutter} (\(\downarrow\)) & \hspace{0.1cm}0.36\hspace{0.1cm}          & \hspace{0.1cm}0.28\hspace{0.1cm}         & \hspace{0.1cm}\underline{0.2}\hspace{0.1cm}             & \hspace{0.1cm}\textbf{0.06}\hspace{0.1cm}    \\ 
\bottomrule
\end{tabular}%
}
\vspace{-1.2em}
\end{table}

\section{Conclusion}
In this study, we introduce a novel DG framework for SAR-ATR, named IRASNet, which integrates CRM and adversarial learning to address the domain gap in radar image recognition. Unlike previous approaches that often suffer from reliance on clutter regions due to inadequate feature-level clutter reduction, IRASNet effectively minimizes the influence of clutter by utilizing positional information of both target and shadow regions. The dual focus allows for more robust feature extraction and enhances model performance across diverse scenarios. Through comprehensive experiments on public SAR benchmark datasets, IRASNet demonstrates superior adaptability to varying background clutter conditions, outperforming existing methods by a significant margin. The proposed framework not only maintains high recognition accuracy under challenging SCR fluctuations but also shows enhanced generalization capabilities when exposed to unseen clutter types, such as grass, trees, and shadows, that were not included in the training data. Crucially, the clutter reduction mechanisms incorporated in IRASNet are shown to significantly improve DG performance and increase class discrimination. By reducing the interference from clutter regions in the feature extraction process, the model better differentiates between classes, leading to higher overall accuracy. Additionally, the introduction of mask encoding effectively integrates positional information into the latent space, which proves vital for accurate target and shadow representation in SAR-ATR. The capability of accurately reflecting spatial details in the latent space is particularly important in SAR-ATR, where precise localization of the target is critical for performance. Furthermore, the results underscore the importance of feature-level clutter reduction and accurate spatial encoding in achieving domain-invariant learning for SAR-ATR tasks. By integrating clutter reduction directly into the feature extraction process, IRASNet achieves state-of-the-art performance.

\bibliographystyle{unsrt}
\bibliography{ref}

\setcounter{equation}{0}
\setcounter{figure}{0}
\setcounter{table}{0}
\setcounter{section}{0}
\makeatletter
\renewcommand{\theequation}{S\arabic{equation}}
\renewcommand{\thefigure}{S\arabic{figure}}
\renewcommand{\thetable}{S\arabic{table}}
\renewcommand{\thesection}{S\arabic{section}}

\begin{center}
    \section*{\textbf{Supplementary Material}}
\end{center}
\addcontentsline{toc}{section}{Supplementary Materials} 

\section{Building Multiple Source Domain}
\label{building_msd}
Feature-level clutter reduction in SAR-ATR problems, having identical distribution between training and testing sets, utilizes data augmentation to enhance clutter reduction performance. Since the characteristics of the target region do not differ between the training and testing phases, data augmentation is not applied to the target region. However, as clutter areas are elements that have to be removed, data augmentation is performed by probabilistically generating clutter variants \cite{cfa}.

\begin{figure}[hbt!]
\centering
\includegraphics[width=0.9\columnwidth]{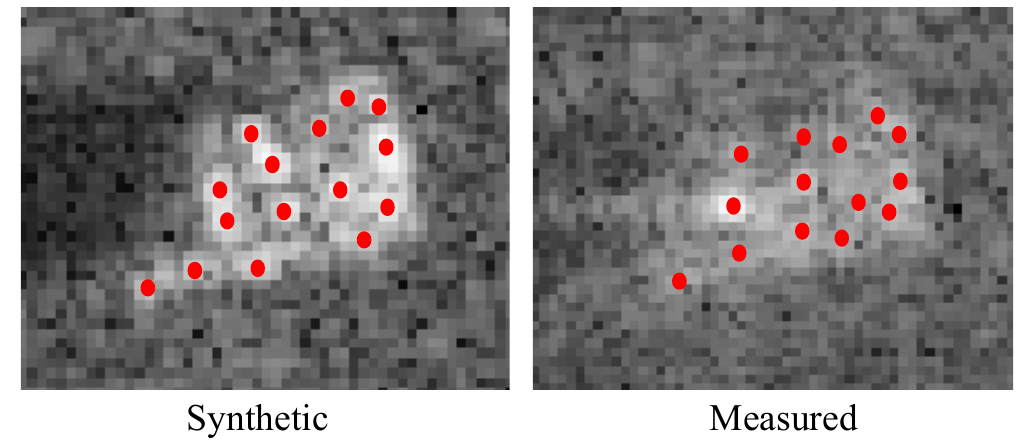}
\caption{Variation in target scattering topology across domains.}
\label{fig:topology}
\end{figure}

DG-ATR problems focus on domain-invariant feature extraction by aligning the distributions of multiple source domains. Therefore, the proposed framework utilizes data augmentation from synthetic datasets based on existing studies \cite{mj}, \cite{inkawhich2021bridging} for robust clutter reduction and constructing multiple source domains. As shown in Fig.~\ref{fig:topology}, the key points of the target region differ across domains. Thus, to realize robust feature-level clutter reduction, additionally reflecting changes in the target signature of the clutter variants is necessary.

We obtain an augmented domain \(D_{Aug}^S\) through augmentation methods applied to the synthetic domain \(D_{Syn}^S\). The augmentation approach from \(D_{Syn}^S\) to \(D_{Aug}^S\) focuses on two types: adding Gaussian noise and altering the signatures of the target, and clutter. To introduce signature changes, \(x^{{Syn}}\) is processed through a Gaussian mixture model to statistically capture the target and clutter regions. In most SAR images, clutter occupies a significant portion. Therefore, the distribution of the clutter region is probabilistically altered through \( \tilde{\mu} = \mu^{Syn} \cdot n_m \) and \( \tilde{\sigma} = \sigma^{Syn} \cdot n_\sigma \), where \( \mu^{Syn} \) and \( \sigma^{Syn} \) represent the mean and standard deviation of \( x^{Syn} \), respectively. We set \( n_m \sim U(1, 1.4) \) and \( n_\sigma \sim U(0.7, 1.3) \) to induce variations in SCR. Additionally, since the target region occupies a relatively small portion, we perform sampling on the histogram distribution to induce fine perturbations in amplitude corresponding to the target region, considering the scattering points and pixel level. The process enhances robustness against RCS variations that occur in real-world scenarios. Further, we construct a cumulative distribution function from the newly generated probability distributions \( \tilde{\mu} \) and \( \tilde{\sigma} \). Then, the augmented version \( \tilde{x} \) is generated through histogram matching \cite{mj}.

Subsequently, to apply Gaussian noise to synthetic data \( \tilde{x} \), we first set the standard deviation \( \sigma_g \) (as a hyper-parameter of Gaussian noise) and then create an augmented version as \( x^{Aug} = \tilde{x} + \mathcal{N} \) \cite{inkawhich2021bridging}. Inkawhich et al. \cite{inkawhich2021bridging} reported that synthetic data lacked background noise compared to measured data. Therefore, adding Gaussian noise is particularly important for effectively estimating the noise. Finally, the values of \( x^{Aug} \) are clipped within the range [0, 1].

The two approaches, utilizing augmentation methods from previous studies \cite{mj}, \cite{inkawhich2021bridging}, primarily address the discovered distribution gaps at the pixel level. However, since the methods alone are insufficient to completely achieve domain invariance at the feature level, this study aims to further reduce the domain gap more precisely by aligning the augmented data features at the feature level.

\begin{algorithm}[t]
\caption{Main learning algorithm of the proposed IRASNet framework}

\DontPrintSemicolon

    \KwInput{
    \begin{itemize}
        \item Batch size \(B\), network structure \(f\).
        \item \textbf{Source Domain:}
        \begin{itemize}
            \item Training SAR samples \(X^{S}=\{X^{Syn}, X^{Aug}\}\).\
            \\ \qquad \qquad \qquad \qquad \qquad \textcolor{blue}{\(\triangleright\) \textit{Data Augmentation}}
            \item Corresponding label \(Y^{S}=\{Y^{Syn}, Y^{Aug}\}\).
            \item Target/shadow mask \(M^{S}_{T/S}=\{M^{Syn}_{T/S}, M^{Aug}_{T/S}\}\).
            \item Assign \(d = 1\) for \(X^{Syn}\) and \(d = 2\) for \(X^{Aug}\).
        \end{itemize}
    \end{itemize}
    }
    \KwOutput{Trained network \(f(\cdot; \theta_F, \theta_D, \theta_Y)\)}
    \KwNT
    
  \, Construct \(f\), composed of \(f_F(\cdot;\theta_F)\) with CRMs, \(f_D(\cdot;\theta_D)\), and \(f_Y(\cdot;\theta_Y)\).
  \\ \, Initialize \(\theta_F, \theta_D, \theta_Y\)

     \While{minibatch \(\mathcal{B} \subset \{1,\cdots,N_B\}, \{X_{(b)}^S\}\subset X^S, \{Y_{(b)}^S\}\subset Y^S, \{M_{T/S, (b)}^{S}\}\subset M_{T/S}^S, d_{(b)}\subset \mathcal{D}\)}
   {
            \While{\(b \in \mathcal{B}\)}
            {
            \small
            \(P(\hat{y}, \hat{d} \mid X_{(b)}^S),F_{TM}, F_{SM} = f(X_{(b)}^S;\theta_F, \theta_D, \theta_Y)\)\;
                \(H_T = g(M_{T, (b)}^{S};\theta_M)\)
                \\ \(H_S = g(M_{S, (b)}^{S};\theta_M)\) \qquad \qquad \textcolor{blue}{\(\triangleright\) \textit{Mask GT encoding}}\;
            
            }
            \small
            \((\theta_F^*, \theta_Y*)=\text{argmin}_{\theta_F,\theta_Y}\mathcal{L}_{cls}(\hat{y},Y_{(b)}^S)-\mathcal{L}_{adv}(\hat{d}, d_{(b)}) \)
            \\ \((\theta_D^*)=\text{argmin}_{\theta_D}\mathcal{L}_{adv}(\hat{d}, d_{(b)})\)
            \\ \((\theta_F^*, \theta_M^*)=\text{argmin}_{\theta_F, \theta_M}\mathcal{L}_{T/S}(F_{TM/SM},H_{T/S})\)\;
   }
   \textbf{return} trained network \(f(\cdot;\theta_F, \theta_D, \theta_Y)\)
\end{algorithm}

\section{Details of the Proposed Framework}
\subsection{Domain-invariant representation learning (DIRL)}
DIRL is a popular DG strategy that aims to learn domain-invariant representations across different domains \cite{Li_2018_CVPR,li2018deep,matsuura2020domain}. Data augmentation methods mitigate the domain gap at a pixel level, whereas adversarial learning for DIRL addresses the domain gap at a feature level using multi-domain datasets with diverse distributions. In SAR-ATR, as acquiring multi-domain datasets is challenging, domain shifts are intentionally induced through data augmentation \cite{inkawhich2021bridging}. Combined with datasets, adversarial learning prevents the feature extractor from distinguishing between different domains, thereby addressing domain shifts at the feature level. The approach enhances generalization performance on both measured and synthetic data by ensuring that domain-invariant features allow the classifier to implement consistent decisions regardless of the domain \cite{matsuura2020domain}. Introducing DG with a domain-invariant feature extraction method based on adversarial learning to SAR-ATR can overcome the limitations posed by the scarcity of measured data and the constraints of data augmentation.

\begin{algorithm}[t]
\caption{Inference algorithm of the proposed IRASNet framework}
\DontPrintSemicolon
  
  \KwInput{
  \\ \, measured SAR sample \(x_i^{Mea}\in X^{Mea}\),
  \\ \, trained network \(f(\cdot;\theta_F^*, \theta_D^*, \theta_Y^*)\)
  }

  \textbf{Network Test}
  \\ \, \(P(\tilde{y} \mid x_i^{Mea}) = f(x_i^{Mea};\theta_F^*, \theta_D^*, \theta_Y^*)\)
  \\ \, \(\tilde{Y} = \text{argmax}_Y P(x_i^{Mea} \mid \theta_F^*, \theta_D^*, \theta_Y^*)\)

   \textbf{return} recognized output \(\tilde{Y}\)

\end{algorithm}

\subsection{Model train and test}
By combining CRM and adversarial learning, the total training objective is described as:
\begin{equation}
    \begin{aligned}
        \underset{\theta_F, \theta_Y}{\min} = & \mathcal{L}_{cls}(\theta_F, \theta_Y) - \lambda \mathcal{L}_{adv}(\theta_F, \theta_D) \\
        & + \lambda \sum_{l=1}^{N_L} (\mathcal{L}_{T,l}(\theta_F) + \mathcal{L}_{S,l}(\theta_F)) \\
        \underset{\theta_D}{\min} = & \mathcal{L}_{adv}(\theta_F, \theta_D)
    \end{aligned}
\label{eq:final_opt}
\end{equation}
, where each loss function is defined as a weighted sum, with weights specified by \(\lambda\), and \(N_L\) represents the number of layers. The total training objective is optimized using an adaptive moment estimation (Adam) optimizer \cite{kingma2014adam}.

Detailed learning and inference algorithms of the proposed IRASNet are presented as Algorithms 1 and 2, respectively. Notably, mask GT encoding is utilized during the training phase to incorporate \(F_{TM}\) and \(F_{SM}\), but it is not used during the inference phase. Thus, the preprocessing step to obtain target and shadow masks is not required, resulting in faster inference speed.

\section{Dataset and Experimental settings}
\begin{figure}[t]
\centering
\includegraphics[width=\columnwidth]{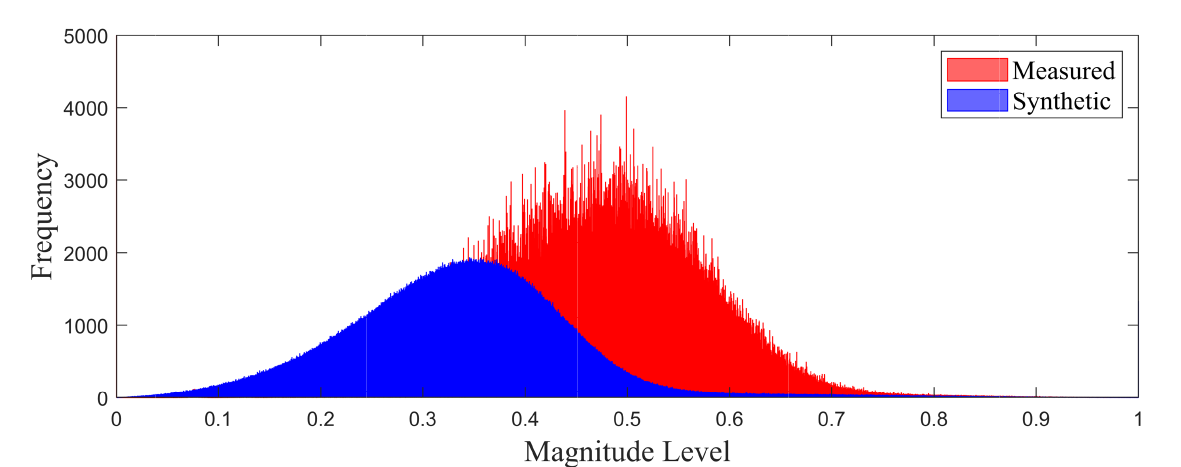}
\caption{Histogram of synthetic and measured data from the SAMPLE dataset. \textit{\textbf{\textcolor{blue}{Blue}}} histogram represents synthetic data, while \textit{\textbf{\textcolor{red}{red}}} histogram represents measured data.}
\label{fig:input_hist}
\end{figure}

\begin{figure}[t]
\centering
\includegraphics[width=0.93\columnwidth]{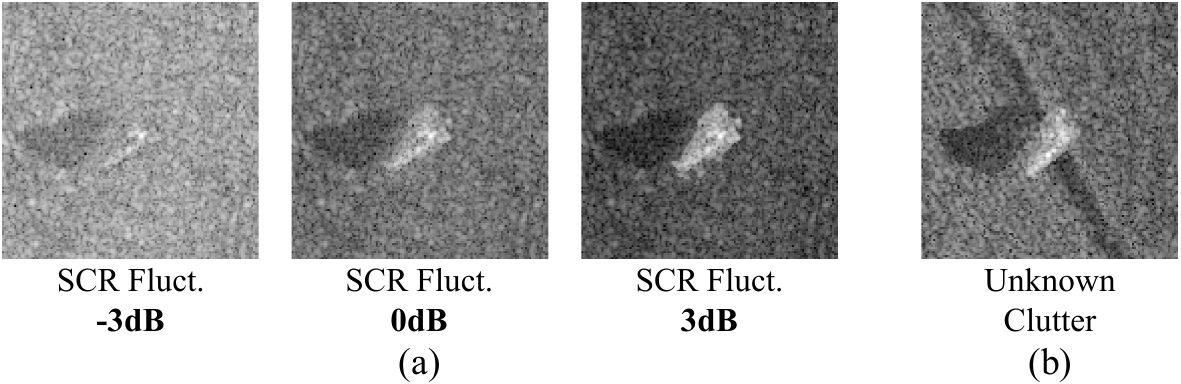}
\caption{(a) Test SAR samples with SCR fluctuations of $-$3, 0, and 3 dB in scenario 3. (b) Test SAR samples for unknown clutter in scenario 4.}
\label{fig:scr_fluct_example}
\end{figure}

The statistical histograms of the synthetic and measured data included in the SAMPLE Dataset are shown in Fig.~\ref{fig:input_hist}, illustrating the distribution differences between the two datasets. Additionally, the SAR images corresponding to Scenario 3 with varying SCR are presented in Fig.~\ref{fig:scr_fluct_example} (a). At $-$3dB, the average brightness of the clutter region was similar to that of the target, while at $+$3dB, the average brightness of the clutter region resembled that of the shadow area. Lastly, the data referred to as "unknown clutter," corresponding to Scenario 4, is shown in Fig.~\ref{fig:scr_fluct_example} (b).


In Section~\ref{building_msd}, the augmented dataset involved a ten-fold augmentation of the synthetic dataset for use in the experiments. The augmented dataset was consistently applied to all comparison models across the experiments to evaluate how well the DL models structurally reduced the domain gap at the feature level.

Additionally, the Mask GT used for training was obtained using the target and shadow segmentation algorithm proposed in a previous study \cite{choi2022fusion}. Since the average brightness and standard deviation differed between the synthetic and measured datasets, hyper-parameter adjustments were necessary for the segmentation algorithm \cite{inkawhich2021bridging}. Therefore, during the training phase of IRASNet, hyper-parameters suitable for the synthetic data were used: minimum target intensity, standard deviation, and shadow threshold of 30, 50, and 0.33, respectively, based on the minimum intensity of the image. During the evaluation phase, hyper-parameters appropriate for the measured dataset were applied: minimum target intensity, standard deviation, and shadow threshold of 35, 50, and 0.45, respectively, based on the minimum intensity of the image. The process was aimed at evaluating clutter reduction performance, and image segmentation was not required to obtain actual ATR outputs. Contrarily, IRASNet did not have hyper-parameters that varied according to the domain, allowing the model to handle changes between different data domains more robustly. Owing to this characteristic, IRASNet delivered consistent performance across various experiments without additional domain-specific adjustments.

IRASNet was trained for 100 epochs, using the Adam optimizer \cite{kingma2014adam} with a learning rate of 0.001 and a batch size of 64. All experiments were conducted on hardware comprising an Intel Xeon Gold 6240 CPU with 128GB of memory and an RTX4090 GPU with 24GB of memory. The software environment included a deep learning workstation running Ubuntu 20.04 and PyTorch 1.11 DL framework.

\section{Ablation study}
In this section, we describe an ablation study conducted based on scenario 2 to analyze the impact of various components of our method on performance. IRASNet comprised five main components: \(F_{TM} \otimes F_{IN}'\), \(F_{SM} \otimes F_{IN}'\), target mask loss, shadow mask loss, which together formed the CRM, and adversarial loss. Herein, \(F_{TM} \otimes F_{IN}'\) and \(F_{SM} \otimes F_{IN}'\) represent \(F_{T}\) and \(F_{S}\), respectively. Not using meant concatenating \(F_{TM}\) and \(F_{SM}\) during the AT phase to refine the features. We evaluated the contribution of each component to performance using two approaches. First, we analyzed the impact of adversarial loss by examining the presence or absence of adversarial loss in the base model and IRASNet. Second, we analyzed the effect of the proposed CRM by evaluating the presence or absence of each component in the CRM. Table~\ref{tab:ablation} shows the overall experimental results obtained through the ablation study.

\begin{table}[t]
\centering
\caption{Ablation experiments of different components in IRASNet.}
\label{tab:ablation}
\renewcommand{\arraystretch}{1.3} 
\resizebox{0.9\columnwidth}{!}{%
\begin{tabular}{cccccccc}
\toprule
\toprule
\textbf{Model} & $F_T$ & $F_S$ & \textbf{$\mathcal{L}_T$} & \textbf{$\mathcal{L}_S$} & \textbf{$\mathcal{L}_{adv}$} & \textbf{Accuracy} &  \\ \midrule
CNN            &                                &                                &                 &                 &                   & 88.62\%           & -                    \\
               &                                &                                &                 &                 & \checkmark        & 92.34\%           & +3.72\%               \\
               &                                &                                & \checkmark      & \checkmark      & \checkmark        & 95.31\%           & +6.69\%               \\
               & \checkmark                     &                                & \checkmark      &                 & \checkmark        & 94.82\%           & +6.20\%               \\
               &                                & \checkmark                     &                 & \checkmark      & \checkmark        & 94.72\%           & +6.10\%               \\
               & \checkmark                     & \checkmark                     &                 &                 & \checkmark        & 95.17\%           & +6.55\%               \\
               & \checkmark                     & \checkmark                     & \checkmark      &                 & \checkmark        & 95.16\%           & +6.54\%               \\
               & \checkmark                     & \checkmark                     &                 & \checkmark      & \checkmark        & 95.24\%           & +6.62\%               \\
               & \checkmark                     & \checkmark                     & \checkmark      & \checkmark      &                   & 92.79\%           & +4.17\%               \\ \midrule
IRASNet        & \checkmark                     & \checkmark                     & \checkmark      & \checkmark      & \checkmark        & \textbf{96.72\%}  & \textbf{+8.10\%}      \\ \bottomrule
\end{tabular}%
}
\end{table}

1) To analyze the impact of adversarial loss, we used a CNN as the baseline model. The results of applying \(L_{adv}\) to the baseline model and removing \(L_{adv}\) from IRASNet indicated that adversarial loss was effective for DG. Meanwhile, we observed from another perspective that using all CRM components in the baseline model significantly improved performance. Without adversarial loss, the recognition accuracy improved to 92.79$\%$, reaching a level comparable to using DG techniques. Although the results were lower than the recognition rate achieved by combining adversarial loss with at least one CRM component, they demonstrated some potential for feature-level clutter reduction.

\begin{figure}[t]
\centering
\includegraphics[width=0.98\columnwidth]{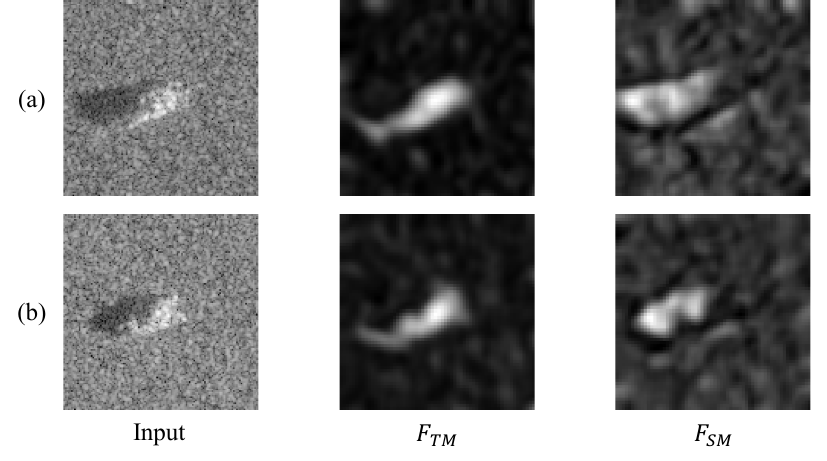}
\caption{Target mask feature (\(F_{TM}\)) and shadow mask feature (\(F_{SM}\)) \\(a) M1. (b) T72.}
\vspace{-1.2em}
\label{fig:ftmsm}
\end{figure}

2) To analyze the impact of CRM components, we used a CNN with adversarial loss as the baseline model. According to Table~\ref{tab:ablation}, introducing \(L_{t}\) and \(L_{s}\) increased performance by approximately 3$\%$ compared to the baseline model, indicating that utilizing the positional information of the target and shadow areas contributed to performance improvement. However, when \(F_{T}\) and \(F_{S}\) were additionally introduced, accuracy increased by approximately 1.4$\%$. When using only \(L_{t}\) and \(L_{s}\), the mask feature was passed to the next layer. The mask feature learned from Mask GT, composed of only 0 and 1, did not secure inter-class discrimination in the feature space even after passing through AT. Herein, Fig.~\ref{fig:ftmsm} shows the visualization of \(F_{tm}\) and \(F_{sm}\). Evidently, \(F_{tm}\) and \(F_{sm}\) omitted the topology that contained scattering points in the target. Therefore, IRASNet, which utilized both \(F_{T}\) and \(F_{S}\), showed improved inter-class discrimination, resulting in higher performance. Meanwhile, the algorithm that introduced only \(F_{T}\) and \(F_{S}\) also improved performance by about 3$\%$ compared to the baseline model. The configuration achieved a certain level of discrimination, but no limitation in clutter reduction was posed as the positional information of the target and shadow was not accurately reflected. Moreover, among the algorithms where positional information was reflected in only one of \(F_{T}\) and \(F_{S}\), using shadow information resulted in higher performance, but no significant difference compared to cases without positional information were observed.

Ultimately, accurately reflecting the positional information of the target and shadow served as a crucial factor for performance improvement, which explained why IRASNet performed better. Finally, using both target and shadow areas simultaneously, with their positional information accurately reflected, increased performance by approximately 2$\%$ compared to using only one area. Thus, the successful utilization of shadow information was demonstrated. Combining any component of CRM with adversarial learning improved performance, proving that integrating clutter reduction and domain-invariant feature learning in the feature space positively impacted DG research in radar image pattern recognition.
\end{document}